\documentclass[sigconf]{acmart}

\usepackage{algorithmic}
\usepackage{float}
\usepackage[linesnumbered,ruled,vlined]{algorithm2e}
\SetKwInput{KwInput}{Input}                
\SetKwInput{KwOutput}{Output}              
\SetKwInput{KwParameter}{Parameter}        

\SetCommentSty{mycommfont}
\usepackage{amsmath}
\usepackage{mathtools, nccmath}
\DeclarePairedDelimiter{\nint}\lfloor\rceil
\usepackage{dsfont}
\usepackage{enumitem}
\usepackage{subfigure}
\usepackage[percent]{overpic}
\usepackage{array}
\usepackage{siunitx}
\usepackage{multirow}
\usepackage{placeins}

\usepackage{adjustbox}
\usepackage{lscape}
\usepackage{threeparttable}
\usepackage{graphicx} 
\usepackage{hyperref}
\usepackage{booktabs}       
\usepackage{amsfonts}       
\usepackage{nicefrac}       
\usepackage{microtype}      
\usepackage{xcolor}         

\newcommand{\boldres}[1]{{\textbf{\textcolor{red}{#1}}}}
\newcommand{\secondres}[1]{{\underline{\textcolor{blue}{#1}}}}

\AtBeginDocument{%
  \providecommand\BibTeX{{%
    \normalfont B\kern-0.5em{\scshape i\kern-0.25em b}\kern-0.8em\TeX}}}
\newcommand{\minus}{\scalebox{0.5}[1.0]{$-$}}

\copyrightyear{2024} 
\acmYear{2024} 
\setcopyright{rightsretained} 
\acmConference[KDD '24]{Proceedings of the 30th ACM SIGKDD Conference on Knowledge Discovery and Data Mining}{August 25--29, 2024}{Barcelona, Spain}
\acmBooktitle{Proceedings of the 30th ACM SIGKDD Conference on Knowledge Discovery and Data Mining (KDD '24), August 25--29, 2024, Barcelona, Spain}\acmDOI{10.1145/3637528.3671673}
\acmISBN{979-8-4007-0490-1/24/08}

\begin{document}

\title{Self-Supervised Learning of Time Series Representation via Diffusion Process and Imputation-Interpolation-Forecasting Mask}

 \author{Zineb Senane$^{*}$}
 \orcid{0009-0001-6451-0136}
 \affiliation{
   \institution{Motherbrain, EQT Group}
   \institution{KTH Royal Institute of Technology}
   \city{Stockholm}
   \country{Sweden}}
 \authornote{Zineb Senane and Lele Cao contributed equally as first authors. For correspondence, please reach out to either of them. The source code and models for reproduction purposes are available at {\color{blue}\url{https://github.com/llcresearch/TSDE}}.}
 \email{senane@kth.se}
 
 \author{Lele Cao$^{*}$}
 \orcid{0000-0002-5680-9031}
 \affiliation{%
  \institution{Motherbrain, EQT Group}
  \city{Stockholm}
  \country{Sweden}
}
\email{caolele@gmail.com}
\email{lele.cao@eqtpartners.com}

 \author{Valentin Leonhard Buchner}
 \orcid{0000-0002-1262-3016}
 \affiliation{%
  \institution{Motherbrain, EQT Group}
  \city{Stockholm}
  \country{Sweden}
}
 \email{vlbuchner@gmail.com}
 
 \author{Yusuke Tashiro}
 \orcid{0009-0000-2659-7122}
 \affiliation{
 \institution{Mitsubishi UFJ Trust Investment Technology Institute}
 \city{Tokyo}
 \country{Japan}
}
\email{yusu.tashi@gmail.com}
 
 \author{Lei You}
 \orcid{0000-0002-4741-0715}
 \affiliation{%
  \institution{Technical University of Denmark}
  \city{Ballerup}
  \country{Denmark}}
 \email{leiyo@dtu.dk}
 
 \author{Pawel Andrzej Herman}
 \orcid{0000-0001-6553-823X}
 \affiliation{%
  \institution{KTH Royal Institute of Technology}
  \city{Stockholm}
  \country{Sweden}}
 \email{paherman@kth.se}
 
 \author{Mats Nordahl}
 \orcid{0009-0000-4347-5928}
 \affiliation{%
  \institution{KTH Royal Institute of Technology}
  \city{Stockholm}
  \country{Sweden}
  }
 \email{mnordahl@kth.se}
 
 \author{Ruibo Tu}
 \orcid{0000-0003-1356-9653}
 \affiliation{%
  \institution{KTH Royal Institute of Technology}
  \city{Stockholm}
  \country{Sweden}
}
 \email{ruibo@kth.se}
 
\author{Vilhelm von Ehrenheim}
\orcid{0000-0002-4210-4989}
\affiliation{%
  \institution{Motherbrain, EQT Group}
  \institution{QA.tech}
  \city{Stockholm}
  \country{Sweden}
}
\email{vonehrenheim@gmail.com}




\renewcommand{\shortauthors}{Zineb Senane and Lele Cao et al.}

\begin{abstract}
Time Series Representation Learning (TSRL) focuses on generating informative representations for various Time Series (TS) modeling tasks. Traditional Self-Supervised Learning (SSL) methods in TSRL fall into four main categories: reconstructive, adversarial, contrastive, and predictive, each with a common challenge of sensitivity to noise and intricate data nuances. Recently, diffusion-based methods have shown advanced generative capabilities. However, they primarily target specific application scenarios like imputation and forecasting, leaving a gap in leveraging diffusion models for generic TSRL. Our work, Time Series Diffusion Embedding (TSDE), bridges this gap as the first diffusion-based SSL TSRL approach. TSDE segments TS data into observed and masked parts using an Imputation-Interpolation-Forecasting (IIF) mask. It applies a trainable embedding function, featuring dual-orthogonal Transformer encoders with a crossover mechanism, to the observed part. We train a reverse diffusion process conditioned on the embeddings, designed to predict noise added to the masked part. Extensive experiments demonstrate TSDE's superiority in imputation, interpolation, forecasting, anomaly detection, classification, and clustering. We also conduct an ablation study, present embedding visualizations, and compare inference speed, further substantiating TSDE's efficiency and validity in learning representations of TS data.
\end{abstract}





\maketitle

\section{Introduction}
\label{sec:intro}
Time Series (TS) data is a sequence of data points collected at regular time intervals.
It is prevalent in various real-world applications, such as understanding human behavioral patterns \cite{challa2022multibranch}, conducting in-depth financial market analyses \cite{cao2022simulation}, predicting meteorological phenomena \cite{karevan2020transductive}, and enhancing healthcare diagnostics \cite{ma2020concare}.
In this work, we focus on Multivariate TS (MTS) data, which refers to a TS with multiple variables or features recorded at each time point, where these variables may have inter-dependencies. 
This is in contrast to Univariate TS (UTS), which only involves a single variable. 
It should be noted that Multiple TS (Multi-TS) differs from MTS as it pertains to the simultaneous monitoring of several UTSs, each operating independently without any interrelation among them.
While this paper primarily concentrates on MTS data, our methodology and insights are also applicable to UTS and Multi-TS, ensuring the versatility and broad applicability of our approach.

To effectively extract and interpret valuable information from intricate raw MTS data, the field of Time Series Representation Learning (TSRL) has become increasingly pivotal. 
TSRL focuses on learning latent representations that encapsulate critical information within the time series, thereby uncovering the intrinsic dynamics of the associated systems or phenomena \cite{meng2023unsupervised}. 
Furthermore, the learned representations are crucial for a variety of downstream applications, such as time series imputation, interpolation, forecasting, classification, clustering and anomaly detection.
TSRL can be conducted in a supervised manner; however, the need for extensive and accurate labeling of vast time series data presents a significant bottleneck, often resulting in inefficiencies and potential inaccuracies. 
Consequently, our focus lies in unsupervised learning techniques, which excel in extracting high-quality MTS representations without the constraints of manual labeling.

Self-Supervised Learning (SSL), a subset of unsupervised learning, has emerged as a highly effective methodology for TSRL. 
SSL utilizes innovative pretext tasks\footnote{A pretext task in SSL is a self-generated learning challenge designed to facilitate the extraction of informative representations for downstream tasks, encompassing various methods such as transformation prediction, masked prediction, instance discrimination, and clustering, tailored to the specific data modality involved \cite{jing2020self,ericsson2022self,zhang2023self}.} to generate supervision signals from unlabeled TS data, thereby facilitating the model's ability to autonomously learn valuable representations without relying on external labels. 
The four main designs of SSL pretext tasks -- reconstructive, adversarial, contrastive, and predictive \cite{liu2021self,zhang2023self,meng2023unsupervised,eldele2023label} -- will be elaborated in Section~\ref{sec:related-work}. 
These designs have demonstrated notable success in addressing TSRL across a diverse range of applications, yet they often struggle with capturing the full complexity of MTS data, particularly in modeling intricate long-term dependencies and handling high-dimensional, noisy datasets.

Due to their advanced generative capabilities, diffusion models  \cite{sohl2015deep,song2019generative,ho2020denoising,song2020improved,song2021maximum,song2021scorebased} have emerged as a promising solution for TS modeling, adept at handling the complexities and long-term dependencies often found in MTS data. 
While these methods have shown success in specific tasks like forecasting \cite{rasul2021autoregressive} and imputation \cite{tashiro2021csdi}, their adoption in SSL~TSRL remains largely unexplored, leaving a gap in the related research literature. 
Our work, Time Series Diffusion Embedding (TSDE), pioneers in this area by integrating conditional diffusion processes with crossover Transformer encoders and introducing an Imputation-Interpolation-Forecasting~(IIF) mask strategy. 
This unique combination allows TSDE to generate versatile representations that are applicable to a wide range of tasks, including imputation, interpolation, forecasting, classification, anomaly detection, and clustering. Our main contributions are:
\begin{itemize}[leftmargin=*]
\item We propose a novel SSL TSRL framework named TSDE, which optimizes a denoising (reverse diffusion) process, conditioned on a learnable MTS embedding function. 
\item We develop dual-orthogonal Transformer encoders integrated with a crossover mechanism, which learns MTS embeddings
by capturing temporal dynamics and feature-specific dependencies.
\item We design a novel SSL pretext task, the IIF masking strategy, which creates pseudo observation masks designed to simulate the typical imputation, interpolation, and forecasting tasks.
\item We experimentally show that TSDE achieves superior performance over existing methods across a wide range of MTS tasks, thereby validating the universality of the learned embeddings.
\end{itemize}

\section{Related work}
\label{sec:related-work}
This research addresses the problem of TSRL using a SSL approach.
Inspired by the taxonomies adopted by \cite{liu2021self,zhang2023self,meng2023unsupervised,eldele2023label}, we structure our review of SSL-based TSRL around four primary methodologies: reconstructive, adversarial, contrastive, and predictive methods.

{\bf Reconstructive methods} focus on minimizing the discrepancy between original and reconstructed MTS data, mostly using an encoder-decoder Neural Network (NN) architecture to emphasize salient features and filter out noise, thereby training the NN to learn meaningful representations \cite{hinton2006reducing}.
Recent mainstream methods in this category predominantly employ Convolutional NN (CNN)~\cite{song2020representation,zhang2021unsupervised}, Recurrent NN (RNN)~\cite{malhotra2017timenet,sagheer2019unsupervised} or Transformer~\cite{chowdhury2022tarnet,zhang2023selfsupervised} as their architectural backbone.
In this category, deep clustering stands out by simultaneously optimizing clustering and reconstruction objectives. It has been implemented through various clustering algorithms, including $k$-means~\cite{xie2016unsupervised,caron2018deep}, Gaussian Mixture Model (GMM)~\cite{jiang2017variational,cao2020simple}, and spectral clustering~\cite{tao2020clustering}.
Reconstructive methods might face limitations in addressing long-term dependencies and adequately representing complex features such as seasonality, trends, and noise in extensive, high-dimensional datasets.

{\bf Adversarial methods} utilize Generative Adversarial Network (GAN) to learn TS representations by differentiating between real and generated data \cite{radford2015unsupervised,mehralian2018rdcgan}. 
These methods often integrate advanced NN architectures or autoregressive models to effectively capture temporal dependencies and generate realistic TS data. 
For instance, TimeGAN \cite{yoon2019time} combines GANs with autoregressive models for temporal dynamics replication, while RGAN \cite{esteban2017real} uses RNN to enhance the realism of generated TS. 
Furthermore, approaches like TimeVAE \cite{desai2021timevae} and DIVERSIFY \cite{lu2022out} innovate in data generation, with the former tailoring outputs to user-specified distributions and latter employing adversarial strategies to maximize distributional diversity in generated TS data.
However, the intricate training process of GANs, potential for mode collapse, and reliance on high-quality datasets are notable drawbacks of adversarial methods, potentially generating inconsistent or abnormal samples \cite{zhang2023self}.

{\bf Contrastive methods} distinguish themselves by optimizing self-discrimination tasks, contrasting positive samples with similar characteristics against negative samples with different ones~\cite{zhu2021an}. 
These methods learn representations by generating augmented views of TS data and leveraging the inherent similarities and variations within the data~\cite{zhang2023self}. 
They include instance-level models~\cite{chen2020simple,chen2021exploring,yang2022timeclr,sun2023test} that treat each sample independently, using data augmentations to form positive and negative pairs. 
Prototype-level models~\cite{li2020prototypical,caron2020unsupervised,zhang2021supporting,meng2023mhccl}, on the other hand, break this independence by clustering semantically similar samples, thereby capturing higher-level semantic information. 
Additionally, temporal-level models~\cite{ijcai2021p324,yue2022ts2vec,yang2022unsupervised,sun2023test} address TS-specific challenges by focusing on scale-invariant representations at individual timestamps, enhancing the understanding of complex temporal dynamics.
However, a common disadvantage across these contrastive methods is their potential to overlook higher-level semantic information, especially when not integrating explicit semantic labels, leading to the generation of potentially misleading negative samples.

{\bf Predictive methods} excel in capturing shared information from TS data by maximizing mutual information from various data slices or augmented views. 
These methods, like TST~\cite{zerveas2021transformer}, wave2vec~\cite{schneider19_interspeech}, CaSS~\cite{chen2022cass} and SAITS~\cite{du2023saits}, focus on predicting future, missing, or contextual information, thereby bypassing the need for full input reconstruction. 
Most recent advancements in this category, such as TEMPO~\cite{cao2023tempo} and TimeGPT~\cite{garza2023timegpt}, leverage LLM (Large Language Model) architectures to effectively decompose and predict complex TS components.
TimeGPT, in particular, stands out as a foundation model specifically for TS forecasting, yet it only treats MTS as Multi-TS.
Lag-Llama~\cite{rasul2023lag}, another notable predictive model, demonstrates strong univariate probabilistic forecasting, trained on a vast corpus of TS data. 
However, the challenge in these methods is their focus on local information, which can limit their capacity to capture long-term dependencies and make them susceptible to noise and outliers, thus affecting their generalization ability.

{\bf Diffusion-based methods} in TS modeling have recently gained traction, leveraging the unique abilities of diffusion models to model the data distribution through a process of injecting and reversing noise~\cite{zhang2023self}. 
These models, like TimeGrad~\cite{rasul2021autoregressive} and CSDI~\cite{tashiro2021csdi}, have been effectively applied to tasks such as forecasting and imputation, employing innovative techniques like RNN-conditioned diffusion and multiple Transformer encoders. 
Recent developments like SSSD~\cite{alcaraz2023diffusionbased} have further evolved the field by integrating structured state space models~\cite{gu2022efficiently} with diffusion processes.
These advancements have showcased the flexibility and potential of diffusion models in handling diverse TS data, with applications ranging from electrical load forecasting with DiffLoad~\cite{wang2023diffload} to predicting spatio-temporal graph evolutions using DiffSTG~\cite{wen2023diffstg}. 
Despite these significant advancements, a notable gap remains in the application of diffusion models for TSRL.
While a recent study~\cite{chen2024deconstructing} demonstrates the efficacy of diffusion models as robust visual representation extractors, their specific adaptation and optimization for TSRL have not been explored.
Our work aims to fill this gap with the innovative TSDE framework, synergistically integrating conditional diffusion processes and crossover Transformer encoders, coupled with an innovative IIF mask strategy, to effectively tackle a wide range of downstream tasks.

\begin{figure*}[!t]
  \begin{center}
    \includegraphics[width=\textwidth]{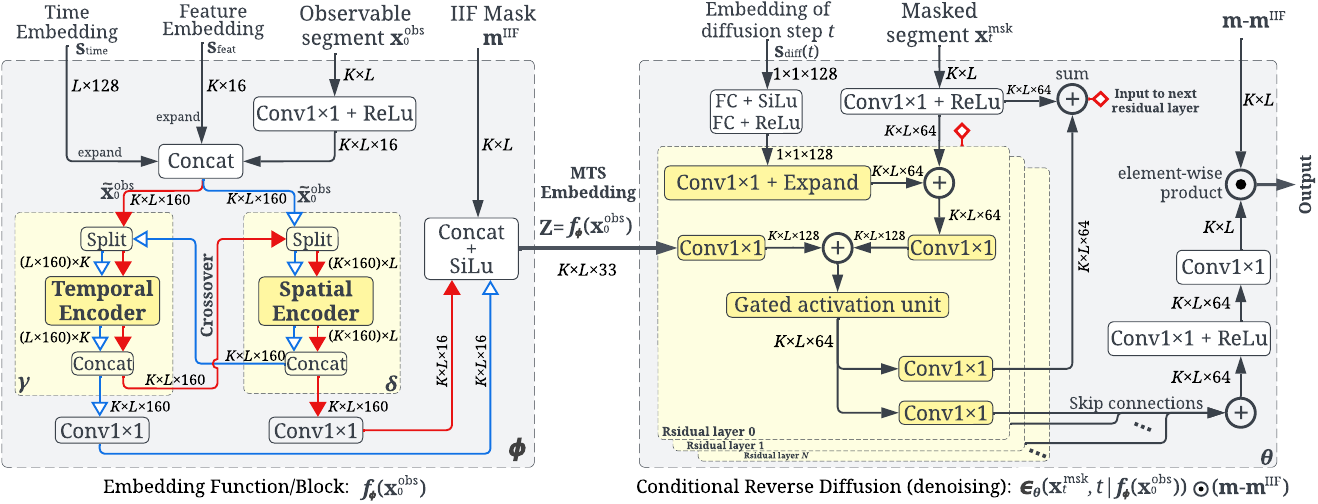}
  \end{center}
  \vspace{-6pt}
  \caption{The TSDE architecture comprises an embedding function (left) and a conditional reverse diffusion block (right): the temporal and spatial encoders are implemented as one-layer Transformer.}
  \label{fig:architecture}
  \vspace{-2pt}
\end{figure*}

\section{The Approach}
The task entails learning general-purpose embeddings for MTS that has $K$ features/variables and $L$ time steps.
Formally, given a multivariate time series $\mathbf{x}$:
\vspace{-2pt}
\begin{equation}
\label{eq:input_X}
  \mathbf{x}=\{x_{1:K,1:L}\}=
  \begin{bmatrix}
    x_{1,1} & x_{1,2} & \ldots & x_{1,L} \\
    x_{2,1} & x_{2,2} & \ldots & x_{2,L} \\
    \vdots & \vdots & \ddots & \vdots \\
    x_{K,1} & x_{K,2} & \ldots & x_{K,L} \\
  \end{bmatrix}\in \mathbb{R}^{K\times L},
\end{equation}
\vspace{-1pt}
we aim to learn a $\boldsymbol{\phi}$-parameterized embedding function $\textit{\textbf{f}}_{\boldsymbol{\phi}}(\cdot)$ that maps the input MTS \textbf{x} to a latent representation $\mathbf{Z}$:
\vspace{-1pt}
\begin{equation}
\label{eq:Z}
  \mathbf{Z}=\{\mathbf{z}_{1:K,1:L}\}=\textit{\textbf{f}}_{\boldsymbol{\phi}}(\mathbf{x})=
  \begin{bmatrix}
    \mathbf{z}_{1,1} & \ldots & \mathbf{z}_{1,L} \\
    \vdots & \ddots & \vdots \\
    \mathbf{z}_{K,1} & \ldots & \mathbf{z}_{K,L} \\
  \end{bmatrix}\in \mathbb{R}^{K\times L\times C},
\end{equation}
where each element $\mathbf{z}_{k,l}\!\in\!\mathbb{R}^C$ represents the embedding vector for the $k$-th feature and $l$-th step, with $C$ denoting the dimensionality of the embedding space.
We propose to learn $\textit{\textbf{f}}_{\boldsymbol{\phi}}$ by leveraging a conditional diffusion process trained in a self-supervised fashion. 

\subsection{Unconditional Diffusion Process}
The unconditional diffusion process assumes a sequence of latent variables $\mathbf{x}_t$ ($t\!\in\!\mathbb{Z}\cap[1,T]$) in the same space as $\mathbf{x}$.
For unification, we will denote $\mathbf{x}$ as $\mathbf{x}_0$ henceforth.
The objective is to approximate the ground-truth MTS distribution $q(\mathbf{x}_0)$ by learning a $\boldsymbol{\theta}$-parameterized model distribution $p_{\boldsymbol{\theta}}(\mathbf{x}_0)$.
The entire process comprises both forward and reverse processes.

\subsubsection{Forward process}
In this process, Gaussian noise is gradually injected to $\mathbf{x}_0$ in $T$ steps until $x_T$ is close enough to a standard Gaussian distribution, which can be expressed as a Markov chain:
\begin{equation}\label{eq:forward_markov}
q(\mathbf{x}_{1:T}|\mathbf{x}_{0})=\textstyle{\prod}_{t=1}^{T} q(\mathbf{x}_{t}|\mathbf{x}_{t-1}),
\end{equation}
where $q(\mathbf{x}_{t}|\mathbf{x}_{t-1})$ is a diffusion transition kernel, and is defined as
\begin{equation}\label{eq:forward_trans_kernel}
    q(\mathbf{x}_{t}|\mathbf{x}_{t-1}):=\mathcal{N}(\mathbf{x}_{t};\sqrt{1-\beta_{t}}\mathbf{x}_{t-1},\beta_{t}\textbf{I}),
\end{equation}
which is a conditional Gaussian distribution 
with a mean of  \newline$\small\smash{\sqrt{1-\beta_{t}}\mathbf{x}_{t-1}}$ and a covariance matrix of $\beta_{t}\textbf{I}$, and $\beta_{t}\!\in\!(0,1)$ indicates the noise level at each diffusion step $t$.
Because of the properties of Gaussian kernels,
we can sample any $\mathbf{x}_{t}$ from $\mathbf{x}_{0}$ directly with
\begin{equation}\label{eq:forward_sampling}
    q(\mathbf{x}_{t}|\mathbf{x}_{0}):=\mathcal{N}(\mathbf{x}_{t};\sqrt{\tilde{\alpha_{t}}}\mathbf{x}_{0},(1-\tilde{\alpha_{t}})\textbf{I})
    , \text{where} \,
    \tilde{\alpha_{t}}:=\textstyle{\prod}_{i=1}^{t} (1-\beta_{i}),
\end{equation}
and $\mathbf{x}_{t}=\sqrt{\tilde{\alpha_{t}}}\mathbf{x}_{0}+\sqrt{1-\tilde{\alpha_{t}}}\boldsymbol{\epsilon}$, and $\boldsymbol{\epsilon}\sim\mathcal{N}(\boldsymbol{0}, \textbf{I})$.

\subsubsection{Reverse process}
This process, modeled by a NN parameterized with $\boldsymbol{\theta}$, recovers $\mathbf{x}_{0}$ by progressively denoising $\mathbf{x}_{T}$:
\begin{equation}\label{eq:eq4}
p_{\boldsymbol{\theta}}(\mathbf{x}_{0:T})=p(\mathbf{x}_{T})\textstyle{\prod}_{t=1}^{T} p_{\boldsymbol{\theta}}(\mathbf{x}_{t-1}|\mathbf{x}_{t}),
\end{equation}
where $p_{\boldsymbol{\theta}}(\mathbf{x}_{t-1}|\mathbf{x}_{t})$ is the reverse transition kernel with a form of 
\begin{equation}\label{eq:uncondition_reverse_kernel}
p_{\boldsymbol{\theta}}(\mathbf{x}_{t-1}|\mathbf{x}_{t}):=\mathcal{N}(\mathbf{x}_{t-1};\boldsymbol{\mu}_{\boldsymbol{\theta}}(\mathbf{x}_{t},t),\boldsymbol{\Sigma}_{\boldsymbol{\theta}}(\mathbf{x}_{t},t)).
\end{equation}
To approximate the reverse transition kernel, Ho et al.~\cite{NEURIPS2020_4c5bcfec} propose the following reparametrization of the mean and variance: 
\begin{align}
  \boldsymbol{\mu}_{\boldsymbol{\theta}}(\mathbf{x}_{t},t)&:=(1-\beta_{t})^{-\!\frac{1}{2}}(\mathbf{x}_{t}-\beta_{t}(1-\tilde{\alpha_{t}})^{-\!\frac{1}{2}}\boldsymbol{\epsilon}_{\boldsymbol{\theta}}(\mathbf{x}_{t}, t)), \label{eq:uncondition_reverse_mean}\\
  \boldsymbol{\Sigma}_{\boldsymbol{\theta}}(\mathbf{x}_{t},t)&:=\boldsymbol{\sigma}_{\boldsymbol{\theta}}(\mathbf{x}_{t},t)\textbf{I}=\sigma_{t}^2\textbf{I}, \label{eq:uncondition_reverse_var}
\end{align}
where $\sigma_{t}^2=\beta_{t}(1-\tilde{\alpha}_{t-1})/(1-\tilde{\alpha}_{t})$ when $t\!>\!1$, otherwise $\sigma_{t}^2=\beta_{1}$;
$\boldsymbol{\epsilon}_{\boldsymbol{\theta}}$ is a trainable network predicting the noise added to input $\mathbf{x}_{t}$ at diffusion step $t$. 
Specifically, $\tilde{\alpha}_{T}\!\approx\!0$ such that $q(\mathbf{x}_{T})\!\approx\!\mathcal{N}(\mathbf{x}_{T};0, \textbf{I})$, thus the starting point of the backward chain is a Gaussian noise.  

\subsection{Imputation-Interpolation-Forecasting Mask}
The reverse process of unconditional diffusion facilitates the generation of MTS from noise. However, our objective is to create general-purpose embeddings for unlabeled MTS, which can be leveraged in many popular downstream tasks such as imputation, interpolation, and forecasting. Consequently, we propose an Imputation-Interpolation-Forecasting (IIF) mask strategy, producing a pseudo observation mask $\small\smash{\mathbf{m}^{\text{IIF}}\!=\!\{m^{\text{IIF}}_{1:K,1:L}\}\!\in\!\{0,1\}^{K\times L}}$ where $\small\smash{m^{\text{IIF}}_{k,l}\!=\!1}$ if $x_{k,l}$ in Equation~\eqref{eq:input_X} is observable, and $\small\smash{m^{\text{IIF}}_{k,l}\!=\!0}$ otherwise.
Algorithm~\ref{algo:masking} details the implementation and combination of imputation, interpolation, and forecasting masks\footnote{The {\it imputation mask} simulates random missing values; the {\it interpolation mask} mimics the MTS interpolation tasks by masking all values at a randomly selected timestamp; and the {\it forecasting mask} assumes all values post a specified timestamp unknown.}.
During training, given any original MTS $\mathbf{x}_0$, we extract the observed ($\small\smash{\mathbf{x}_0^{\text{obs}}}$) and masked ($\small\smash{\mathbf{x}_0^{\text{msk}}}$) segments by
\begin{equation}\label{eq:obs_msk_parts}
\mathbf{x}_0^{\text{obs}} := \mathbf{x}_{0}\odot\mathbf{m}^{\text{IIF}}
\;\; \text{and} \;\;\;
\mathbf{x}_0^{\text{msk}} := \mathbf{x}_{0}\odot(\mathbf{m}-\mathbf{m}^{\text{IIF}})\, ,
\end{equation}
where $\odot$ represents element-wise product; 
and $\mathbf{m}\!=\!\{m_{1:K,1:L}\}\!\in\!\{0,1\}^{K\times L}$ is a mask with zeros indicating originally missing values in $\mathbf{x}_0$.
We now reformulate our self-supervised learning objective to generate the masked version of MTS, denoted as $\mathbf{x}_0^{\text{msk}}$, from a {\bf corrupted input} $\mathbf{x}_t^{\text{msk}}$, through a diffusion process, {\bf conditioned on the embedding of the observed MTS} $\mathbf{x}_0^{\text{obs}}$, i.e.,~$\textit{\textbf{f}}_{\boldsymbol{\phi}}(\mathbf{x}_0^{\text{obs}})$. Both the diffusion process (parameterized by $\boldsymbol{\theta}$) and the embedding function (parameterized by $\boldsymbol{\phi}$) are approximated with a trainable NN.


\subsection{Conditional Reverse Diffusion Process}
\label{sec:cond_rev_diff_proc}
Our conditional diffusion process estimates the ground-truth conditional probability
$q(\textbf{x}_{0}^{\text{msk}}|\textit{\textbf{f}}_{\boldsymbol{\phi}}(\textbf{x}_{0}^{\text{obs}}))$ by re-formulating \eqref{eq:eq4} as
\begin{equation}\label{eq:condition_reverse_process}
\begin{gathered}
p_{\boldsymbol{\theta}}(\textbf{x}_{0:T}^{\text{msk}}|\textit{\textbf{f}}_{\boldsymbol{\phi}}(\textbf{x}_{0}^{\text{obs}}))\!:=\!
p(\textbf{x}_{T}^{\text{msk}})\textstyle{\prod}_{t=1}^{T} p_{\boldsymbol{\theta}}(\textbf{x}_{t-1}^{\text{msk}}|\textbf{x}_{t}^{\text{msk}}\!,\textit{\textbf{f}}_{\boldsymbol{\phi}}(\textbf{x}_{0}^{\text{obs}})\!).
\end{gathered}
\end{equation}
Similar to \eqref{eq:uncondition_reverse_kernel}, the reverse kernel $p_{\boldsymbol{\theta}}(\textbf{x}_{t-1}^{\text{msk}}|\textbf{x}_{t}^{\text{msk}}\!,\textit{\textbf{f}}_{\boldsymbol{\phi}}(\textbf{x}_{0}^{\text{obs}})):=$
\begin{equation}\label{eq:condition_reverse_kernel}
\begin{gathered}
\mathcal{N}(\textbf{x}_{t-1}^{\text{msk}};\boldsymbol{\mu}_{\boldsymbol{\theta}}(\textbf{x}_{t}^{\text{msk}},t,\textit{\textbf{f}}_{\boldsymbol{\phi}}(\textbf{x}_{0}^{\text{obs}})),\boldsymbol{\Sigma}_{\boldsymbol{\theta}}(\textbf{x}_{t}^{\text{msk}},t,\textit{\textbf{f}}_{\boldsymbol{\phi}}(\textbf{x}_{0}^{\text{obs}}))).
\end{gathered}
\end{equation}
According to DDPM~\cite{NEURIPS2020_4c5bcfec}, the variance $\boldsymbol{\Sigma}_{\boldsymbol{\theta}}(\textbf{x}_{t}^{\text{msk}}\!,t,\textit{\textbf{f}}_{\boldsymbol{\phi}}(\textbf{x}_{0}^{\text{obs}}))$ can be formulated in the same way as \eqref{eq:uncondition_reverse_var}, i.e.,~$\boldsymbol{\sigma}_{\boldsymbol{\theta}}(\textbf{x}_{t}^{\text{msk}},\!t,\textit{\textbf{f}}_{\boldsymbol{\phi}}(\textbf{x}_{0}^{\text{obs}}))\textbf{I}=\sigma_{t}^2\textbf{I}$.
Similar to Equation~\eqref{eq:uncondition_reverse_mean}, the conditional mean $\boldsymbol{\mu}_{\boldsymbol{\theta}}(\textbf{x}_{t}^{\text{msk}}\!,t,\textit{\textbf{f}}_{\boldsymbol{\phi}}(\textbf{x}_{0}^{\text{obs}}))\!\!:=$
\begin{equation}\label{eq:condition_reverse_var}
(1-\beta_{t})^{-\!\frac{1}{2}}(\textbf{x}_{t}^{\text{msk}}-\beta_{t}(1-\tilde{\alpha}_{t})^{-\!\frac{1}{2}}\boldsymbol{\epsilon}_{\boldsymbol{\theta}}(\textbf{x}_{t}^{\text{msk}},t \vert \textit{\textbf{f}}_{\boldsymbol{\phi}}(\textbf{x}_{0}^{\text{obs}}))).
\end{equation}

\subsection{Training Loss and Procedure}
It has been shown in \cite{NEURIPS2020_4c5bcfec} that the reverse process of unconditional diffusion can be trained by minimizing the following loss:
\begin{equation}\label{eq:obj_uncondition}
    \mathcal{L}(\boldsymbol{\theta}):= \mathbb{E}_{\textbf{x}_{0} \sim q(\textbf{x}_{0}),\boldsymbol{\epsilon} \sim \mathcal{N}(\boldsymbol{0},\textbf{I}),t} \|\boldsymbol{\epsilon}-\boldsymbol{\epsilon}_{\boldsymbol{\theta}}(\textbf{x}_{t},t))\|_2^2\;.
\end{equation}
Inspired by \cite{tashiro2021csdi}, we replace the noise prediction NN $\boldsymbol{\epsilon}_{\boldsymbol{\theta}}(\textbf{x}_{t},t)$ with the conditioned version $\boldsymbol{\epsilon}_{\boldsymbol{\theta}}(\textbf{x}_{t}^{\text{msk}},t \vert \textit{\textbf{f}}_{\boldsymbol{\phi}}(\textbf{x}_{0}^{\text{obs}})$ in \eqref{eq:obj_uncondition}, obtaining
\begin{equation}\label{eq:obj_condition}
    \mathcal{L}(\boldsymbol{\theta},\boldsymbol{\phi}):= \mathbb{E}_{\textbf{x}_{0} \sim q(\textbf{x}_{0}),\boldsymbol{\epsilon} \sim \mathcal{N}(\boldsymbol{0},\textbf{I}),t} \|\boldsymbol{\epsilon}-\boldsymbol{\epsilon}_{\boldsymbol{\theta}}(\textbf{x}_{t}^{\text{msk}},t \vert \textit{\textbf{f}}_{\boldsymbol{\phi}}(\textbf{x}_{0}^{\text{obs}}))\|_2^2\;.
\end{equation}
Given the focus of training is solely on predicting the noise at the non-missing and masked locations, we actually minimize $\widetilde{\mathcal{L}}(\boldsymbol{\theta},\boldsymbol{\phi}):=$ \begin{equation}\label{eq:obj_masked_condition}
\mathbb{E}_{\textbf{x}_{0}\!\sim q(\textbf{x}_{0}),\boldsymbol{\epsilon}\sim \mathcal{N}(\boldsymbol{0},\textbf{I}),t} \|(\boldsymbol{\epsilon}\!-\!\boldsymbol{\epsilon}_{\boldsymbol{\theta}}(\textbf{x}_{t}^{\text{msk}}\!,t \vert \textit{\textbf{f}}_{\boldsymbol{\phi}}(\textbf{x}_{0}^{\text{obs}})))\!\odot\!(\mathbf{m}\!-\!\mathbf{m}^{\text{IIF}})\|_2^2.
\end{equation}
The self-supervised and mini-batch training procedure, detailed in Algorithm~\ref{algo:training}, essentially attempts to solve $\min_{\boldsymbol{\theta},\boldsymbol{\phi}} \widetilde{\mathcal{L}}(\boldsymbol{\theta},\boldsymbol{\phi})$.
In each iteration $i$ of the training process, a random diffusion step $t$ is chosen, at which point the denoising operation is applied.

\subsection{Embedding Function}
\label{sec:embedding_function}
The left part of Figure~\ref{fig:architecture} illustrates the architectural design of the embedding function $\small\smash{\textit{\textbf{f}}_{\boldsymbol{\phi}}(\textbf{x}_{0}^{\text{obs}})}$. 
This figure highlights that the function not only processes the input $\textbf{x}_{0}^{\text{obs}}$, but also incorporates additional side information (namely, time embedding $\textbf{s}_{\text{time}}(l)$, feature embedding $\textbf{s}_{\text{feat}}(k)$, and the mask $\mathbf{m}^{\text{IIF}}$) into its computations.
Consequently, the notation $\small\smash{\textit{\textbf{f}}_{\boldsymbol{\phi}}(\textbf{x}_{0}^{\text{obs}})}$ is succinctly used to represent the more extensive formulation $\small\smash{\textit{\textbf{f}}_{\boldsymbol{\phi}}(\textbf{x}_{0}^{\text{obs}}, \textbf{s}_{\text{time}}, \textbf{s}_{\text{feat}}, \mathbf{m}^{\text{IIF}})}$, which accounts for all the inputs processed by the function.
To obtain 128-dimensional $\textbf{s}_{\text{time}}(l)$, we largely follow \cite{zuo2020transformer,tashiro2021csdi}:
\vspace{-2pt}
\begin{equation}\label{eq:temp_embedding}
    \textbf{s}_{\text{time}}(l)=\left(
    \sin\frac{l}{\tau^{\frac{0}{64}}}
    ,\ldots,
    \sin\frac{l}{\tau^{\frac{63}{64}}},
    \cos\frac{l}{\tau^{\frac{0}{64}}}
    ,\ldots,
    \cos\frac{l}{\tau^{\frac{63}{64}}}
    \right),
\end{equation}
where $\tau$=10,000 and $l\!\in\!\mathbb{Z}\cap[1,L]$.
For $\textbf{s}_{\text{feat}}(k)$, a 16-dimensional feature embedding is obtained by utilizing the categorical feature embedding layer available in PyTorch.
The observable segment $\textbf{x}_{0}^{\text{obs}}$ undergoes a nonlinear transformation and is then concatenated with time and feature embeddings, resulting in $\tilde{\textbf{x}}_{0}^{\text{obs}}\in\mathbb{R}^{K\times L \times 160}$:
\vspace{-2pt}
\begin{equation}\label{eq:tilde_x}
    \tilde{\textbf{x}}_{0}^{\text{obs}} = Concat(ReLu(Conv(\textbf{x}_{0}^{\text{obs}})),\textbf{s}_{\text{time}},\textbf{s}_{\text{feat}}),
\end{equation}
where $Concat(\cdot)$, $ReLu(\cdot)$ and $Conv(\cdot)$ represent concatenation, ReLu activation, and 1$\times$1 convolution operation \cite{lin2013network} respectively.

To accurately capture the inherent temporal dependencies and feature correlations in MTS data, thereby enabling clearer data interpretation and a customizable, modular design, we devise separate temporal and feature embedding functions: $\textit{\textbf{g}}_{\gamma}(\tilde{\textbf{x}}_{0}^{\text{obs}})$ and $\textit{\textbf{h}}_{\delta}(\tilde{\textbf{x}}_{0}^{\text{obs}})$, parameterized by $\gamma$ and $\delta$ respectively.
Inspired by \cite{tashiro2021csdi}, both the temporal $\textit{\textbf{g}}_{\gamma}(\cdot)$ and feature $\textit{\textbf{g}}_{\delta}(\cdot)$ encoders are simply implemented as a one-layer Transformer encoder that takes an input tensor shaped $K\!\times\!L\!\times\!160$, as shown in Figure~\ref{fig:architecture}. 
Specifically, the temporal encoder operates on tensors shaped $1\!\times\!L\!\times\!160$, representing a feature across all timestamps;
and the feature encoder handles tensors shaped $K\!\times\!1\!\times\!160$, representing a feature vector corresponding to a time stamp.

To integrate temporal and feature embeddings in varying orders without adding to the model's trainable parameters, we have developed a crossover mechanism.
This mechanism is depicted by the red and blue arrows in Figure~\ref{fig:architecture}.
It facilitates the generation of $\textit{\textbf{g}}_{\gamma}(\textit{\textbf{h}}_{\delta}(\tilde{\textbf{x}}_{0}^{\text{obs}}))$ and $\textit{\textbf{h}}_{\delta}(\textit{\textbf{g}}_{\gamma}(\tilde{\textbf{x}}_{0}^{\text{obs}}))$, which are subsequently transformed and concatenated along with $\mathbf{m}^{\text{IIF}}$, resulting in the final embedding $\mathbf{Z}=\textit{\textbf{f}}_{\boldsymbol{\phi}}(\textbf{x}_{0}^{\text{obs}}):=$ 
\vspace{-4pt}
\begin{equation}
\label{eq:final_embedding}
\!\!\!\!SiLu\!\left(Concat\!\left(
Conv(\textit{\textbf{g}}_{\gamma}\!(\textit{\textbf{h}}_{\delta}(\tilde{\textbf{x}}_{0}^{\text{obs}})\!)\!),Conv(\textit{\textbf{h}}_{\delta}(\textit{\textbf{g}}_{\gamma}\!(\tilde{\textbf{x}}_{0}^{\text{obs}})\!)\!),\mathbf{m}^{\text{IIF}}\right)\!\right)\!,
\end{equation}
where $SiLu(\cdot)$ is the Sigmoid-weighted Linear Unit (SiLU) activation function \cite{elfwing2018sigmoid}.
Once the model is trained, the embedding for any MTS $\textbf{x}_{0}$ is computed following Equations \eqref{eq:tilde_x} and \eqref{eq:final_embedding}, where $\textbf{x}_{0}^{\text{obs}}$ and $\mathbf{m}^{\text{IIF}}$ are substituted with $\textbf{x}_{0}$ and $\mathbf{m}$, respectively. 

\setlength{\textfloatsep}{10pt plus 1.0pt minus 2.0pt}
\begin{algorithm}[t!]
\footnotesize
\DontPrintSemicolon
\KwInput{Mask $\mathbf{m}\!=\!\{m_{1:K,1:L}\}\!\in\!\{0,1\}^{K\times L}$ indicating the missing values in $\mathbf{x}_0$}
\KwOutput{A pseudo observation mask $\mathbf{m}^{\text{IIF}}\in\!\{0,1\}^{K\times L}$}



$r \gets$ random value from the range of [0.1, 0.9]; \tcp*{\scriptsize imputation mask ratio}

$N \gets \sum_{k=1}^K \sum_{l=1}^L m_{k,l}$; \tcp*{\scriptsize total number of observed values}


$\mathbf{m}^{\text{IIF}} \gets \mathbf{m}$ and randomly set $\nint{N\times r}$ 1s to 0; \tcp*{\scriptsize apply imputation mask}


Sample a probability $p$ uniformly from the range of [0, 1];

\uIf{$1/3<p<2/3$}{
    $l' \gets$ uniformly sample a time step from $\mathbb{Z}\cap[1,L]$; \\
    $\mathbf{m}^{\text{IIF}}[ : ,l'] \gets 0$; \tcp*{\scriptsize mix with interpolation mask}
}
\uElseIf{$p>=2/3$}{
    $l' \gets$ uniformly sample a time window length from $\mathbb{Z}\cap[1,\nint{\frac{L}{3}}]$; \\
    $\mathbf{m}^{\text{IIF}}[:, -l'\!\!:] \gets 0$; \tcp*{\scriptsize mix with forecasting mask}
}

\textbf{return} $\mathbf{m}^{\text{IIF}}$;
\caption{Imputation-Interpolation-Forecasting Mask}
\label{algo:masking}
\end{algorithm}

\subsection{The Overall Architecture}
\label{sec:overall_architecture}
Figure~\ref{fig:architecture} provides a comprehensive depiction of the various components within the TSDE architecture. 
The process begins by applying the IIF mask $\small\smash{\mathbf{m}^{\text{IIF}}}$ to partition the input MTS into observable ($\small\smash{\textbf{x}_{0}^{\text{obs}}}$) and masked ($\small\smash{\textbf{x}_{0}^{\text{msk}}}$) segments. 
The entire architecture primarily consists of two key elements: 
(1) an embedding function $\small\smash{\textit{\textbf{f}}_{\boldsymbol{\phi}}(\textbf{x}_{0}^{\text{obs}})}$ thoroughly introduced in Section~\ref{sec:embedding_function}; and
(2) a conditional reverse diffusion module, illustrated on the right side of Figure~\ref{fig:architecture}.

The conditional reverse diffusion, introduced in Section~\ref{sec:cond_rev_diff_proc}, functions as a noise predictor, effectively implementing $\small\smash{\boldsymbol{\epsilon}_{\boldsymbol{\theta}}(\textbf{x}_{t}^{\text{msk}},t \vert \textit{\textbf{f}}_{\boldsymbol{\phi}}(\textbf{x}_{0}^{\text{obs}}))}$.
During the $i$-th training step, as outlined in Algorithm~\ref{algo:training}, the sampled diffusion step $t$ is first transformed into a 128-dimensional vector, denoted as $\textbf{s}_{\text{diff}}(t):=$
\vspace{-4pt}
\begin{equation}
\label{eq:diffusion_step_embedding}
\left(
    \sin(10^{\frac{0\cdot 4}{63}}t)
    ,\ldots,
    \sin(10^{\frac{63\cdot 4}{63}}t),
    \cos(10^{\frac{0\cdot 4}{63}}t)
    ,\ldots,
    \cos(10^{\frac{63\cdot 4}{63}}t)
\right).
\vspace{-3pt}
\end{equation}
Subsequently, the MTS embedding $\mathbf{Z}$, along with $\textbf{s}_{\text{diff}}(t)$ and $\small\smash{\textbf{x}_{0}^{\text{msk}}}$, are input into a residual block composed of $N$ residual layers. 
The outputs of these layers are aggregated (summation), processed through some transformations, and combined with $\small\smash{\textbf{x}_{t}^{\text{msk}}}$.
This results in $\small\smash{\boldsymbol{\epsilon}_{\boldsymbol{\theta}}(\textbf{x}_{t}^{\text{msk}}\!,t \vert \textit{\textbf{f}}_{\boldsymbol{\phi}}(\textbf{x}_{0}^{\text{obs}}))\!\odot\!(\mathbf{m}\!-\!\mathbf{m}^{\text{IIF}})}$, which is then utilized to compute the loss $\small\smash{\widetilde{\mathcal{L}}(\boldsymbol{\theta},\boldsymbol{\phi})}$, as formulated in Equation~\eqref{eq:obj_masked_condition}.

\vspace{-2pt}
\subsection{Downstream Tasks and Model Efficiency}
\label{sec:overall_architecture}
The trained model can be utilized in two scenarios: (1) the embedding function, as a standalone component, can be used to generate comprehensive MTS representations, which are suitable for various downstream applications including anomaly detection, clustering, and classification as demonstrated in Section~\ref{sec:exp-anomaly-detection}, \ref{sec:exp-classification}, and \ref{sec:exp-clustering}, respectively. 
(2) When combined with the trained conditional reverse diffusion process, the model is capable of predicting missing values (for imputation and interpolation) as well as future values (for forecasting) in MTS data.
In the second scenario, a notable increase in speed can be achieved compared to the existing diffusion-based methods such as those in \cite{tashiro2021csdi,rasul2021autoregressive}. 
This efficiency, confirmed in Section~\ref{sec:exp-inf-efficiency}, stems from simplifying the conditional reverse diffusion (the right block of Figure~\ref{fig:architecture}, i.e.,~$\boldsymbol{\epsilon}_{\boldsymbol{\theta}}$) to use only Conv1$\times$1 operators.
This streamlining significantly accelerates the $T$=50 steps reverse diffusion process.
\begin{algorithm}[t!]
\footnotesize
\DontPrintSemicolon
\KwInput{Ground-truth MTS data distribution $q(\textbf{x}_{0})$, noise scheduler $\{\tilde{\alpha}_{t}\}$, the denoising and embedding functions (approx. by NN): $\boldsymbol{\epsilon}_{\boldsymbol{\theta}}(\cdot)$ and $\textit{\textbf{f}}_{\boldsymbol{\phi}}(\cdot)$}
\KwOutput{The trained NN parameters $\boldsymbol{\theta}$ and $\boldsymbol{\phi}$}

\KwParameter{The total number of training iterations $N_{\text{train}}$ and learning rate $\tau$}

\For{$(i = 1; i \leq N_{\text{train}}; i++)$}{

    Sample a diffusion step $t\!\sim\!\text{Uniform}(\{1,\ldots, T\})$ and a MTS $\textbf{x}_{0}\!\sim\!q(\textbf{x}_{0})$;

    Obtain IIF Masking $\mathbf{m}^{\text{IIF}}$ by following Algorithm~\ref{algo:masking};

    Obtain the observed ($\small\smash{\mathbf{x}_0^{\text{obs}}}$) and masked ($\small\smash{\mathbf{x}_0^{\text{msk}}}$) parts using Equation~\eqref{eq:obs_msk_parts};

    Sample a noise matrix $\boldsymbol{\epsilon} \sim \mathcal{N}(\mathbf{0},\textbf{I})$ that has the same shape as $\small\smash{\textbf{x}_{0}^{\text{msk}}}$;

    Compute $\textbf{x}_{t}^{\text{msk}} \gets \sqrt{\tilde{\alpha}_{t}}\textbf{x}_{0}^{\text{msk}} + \sqrt{1-\tilde{\alpha}_{t}}\boldsymbol{\epsilon}$;

    Compute loss $\widetilde{\mathcal{L}}\!:=\!\|(\boldsymbol{\epsilon}\!-\!\boldsymbol{\epsilon}_{\boldsymbol{\theta}}(\textbf{x}_{t}^{\text{msk}},t \vert \textit{\textbf{f}}_{\boldsymbol{\phi}}(\textbf{x}_{0}^{\text{obs}})))\odot(\mathbf{m}\!-\!\mathbf{m}^{\text{IIF}})\|_2^2$, cf.~\eqref{eq:obj_masked_condition};

    $\boldsymbol{\theta}:=\boldsymbol{\theta}-\tau\frac{\partial\widetilde{\mathcal{L}}}{\partial\boldsymbol{\boldsymbol{\theta}}}\;$ and
    $\;\boldsymbol{\phi}:=\boldsymbol{\phi}-\tau\frac{\partial\widetilde{\mathcal{L}}}{\partial\boldsymbol{\boldsymbol{\phi}}}$;
}

\textbf{return} $\boldsymbol{\theta}$ and $\boldsymbol{\phi}$;
\caption{TSDE Training Procedure}
\label{algo:training}
\end{algorithm}

\section{Experiments}
Our evaluation of the TSDE framework includes thorough experiments across six tasks (imputation, interpolation, forecasting, anomaly detection, classification, and clustering) accompanied by additional analyses on inference efficiency, ablation study, and embedding visualization. For experiment details, dataset specifications, hyperparameters, and metric formulas, refer to Appendix~\ref{appendix}.
\begin{table*}[t!]
\caption[Results]{\small Probabilistic MTS imputation and interpolation benchmarking results, featuring TSDE's pretraining-only and task-specific finetuned (TSDE+ft) models against established baselines. We present mean and standard deviation (SD) from three iterations, with baseline results primarily derived or reproduced according to \cite{tashiro2021csdi}.}
\vspace{-4pt}
\label{tab_imputation_interpolation_results}
\centering
\footnotesize
\tabcolsep=0.05cm
\scalebox{0.955}{
\begin{tabular}{l l  c c c  c c c  c c c  c c c }

 \multirow{3}{*}{  } & \multirow{3}{*}{Models} & \multicolumn{9}{c}{PhysioNet} & \multicolumn{3}{c}{PM2.5} \\
\cmidrule{3-14} 
 && \multicolumn{3}{c}{10\% masking ratio} & \multicolumn{3}{c}{50\% masking ratio} & \multicolumn{3}{c}{90\% masking ratio} & \multicolumn{3}{c}{} \\
 
 & & \multicolumn{1}{c}{CRPS}  & \multicolumn{1}{c}{MAE} & \multicolumn{1}{c}{RMSE} & \multicolumn{1}{c}{CRPS}  & \multicolumn{1}{c}{MAE} & \multicolumn{1}{c}{RMSE} & \multicolumn{1}{c}{CRPS}  & \multicolumn{1}{c}{MAE} & \multicolumn{1}{c}{RMSE} & \multicolumn{1}{c}{CRPS}  & \multicolumn{1}{c}{MAE} & \multicolumn{1}{c}{RMSE}   \\  
\midrule
 \multirow{7}{*}{\rotatebox{90}{Imputation}}  & {BRITS \cite{NEURIPS2018_734e6bfc}} & {-} & {0.284(0.001)} & {0.619(0.022)} & {-} & {0.368(0.002)} & {0.693(0.023)} & {-} & {0.517(0.002)} & {0.836(0.015)} & {-} & {14.11(0.26)} & {24.47(0.73)} \\
 & {V-RIN \cite{9370004}} & {0.808(0.008)} & {0.271(0.001)} & {0.628(0.025)} & {0.831(0.005)} & {0.365(0.002)} & {0.693(0.022)}  & {0.922(0.003)} & {0.606(0.006)} & {0.928(0.013)} & {0.526(0.025)} & {25.4(0.062)} & {40.11(1.14)} \\

 & {GP-VAE \cite{pmlr-v108-fortuin20a}} & {0.558(0.001)*} & {0.449(0.002)*} & {0.739(0.001)*} & {0.642(0.003)*} & {0.566(0.004)*} & {0.898(0.005)*} & {0.748(0.002)*} & {0.690(0.002)*} & {1.008(0.002)*} & {0.397(0.009)} & {-} & {-} \\
 & {unc. CSDI \cite{tashiro2021csdi}} & {0.360(0.007)} & {0.326(0.008)} & {0.621(0.020)} & {0.458(0.008)} & {0.417(0.010)} & {0.734(0.024)} & {0.671(0.007)} & {0.625(0.010)} & {0.940(0.018)} & {0.135(0.001)} & {12.13(0.07)} & {22.58(0.23)} \\
 & {CSDI \cite{tashiro2021csdi}} & {0.238(0.001)} & {0.217(0.001)} & {0.498(0.020)} & {0.330(0.002)} & {0.301(0.002)} & {\textbf{0.614(0.017)}} & {0.522(0.002)} & {0.481(0.003)} & {0.803(0.012)} & \underline{0.108(0.001)} & {\textbf{9.60(0.04)}} & \underline{19.30(0.13)} \\
 & {\textbf{TSDE}} & {\textbf{0.226(0.002)}} & {\textbf{0.208(0.001)}} & {\textbf{0.446(0.003)}} & {\textbf{0.316(0.000)}} & {\textbf{0.290(0.000)}} & \underline{0.641(0.007)} & {\textbf{0.488(0.001)}} & {\textbf{0.450(0.001)}} & {\textbf{0.801(0.001)}} & {0.13(0.001)} & {11.41(0.60)} & {27.02(2.91)} \\
  & {\textbf{TSDE+ft}} & \underline{0.230(0.001)} & \underline{0.211(0.001)} & \underline{0.4718(0.013)} & \underline{0.318(0.001)} & \underline{0.292(0.001)} & {0.644(0.001)} & \underline{0.490(0.001)} & \underline{0.452(0.001)} & \underline{0.803(0.001)} & {\textbf{0.107 (0.000)}} & \underline{9.71(0.04)} & {\textbf{18.76(0.02)}} \\
 
\midrule

\multirow{4}{*}{\rotatebox{90}{Interpolation}} & {Latent ODE \cite{NEURIPS2019_42a6845a}} & {0.700(0.002)} & {0.522(0.002)} & {0.799(0.012)} & {0.676(0.003)} & {0.506(0.003)} & {0.783(0.012)} & {0.761(0.010)} & {0.578(0.009)} & {0.865(0.017)} & \multicolumn{3}{c}{* Results reproduced using GP-VAE}\\
 & {mTANs \cite{shukla2021multitime}} & {0.526(0.004)} & {0.389(0.003)} & {0.749(0.037)} & {0.567(0.003)} & {0.422(0.003)} & {0.721(0.014)} &{0.689(0.015)} & {0.533(0.005)} & {0.836(0.018)} & \multicolumn{3}{c}{original implementation available at}\\
 & {CSDI \cite{tashiro2021csdi}} & {0.380(0.002)} & {0.362(0.001)} & {0.722(0.043)} & \underline{0.418(0.001)} & {0.394(0.002)} & {0.700(0.013)} & \underline{0.556(0.003)} & \underline{0.518(0.003)} & {0.839(0.009)} & \multicolumn{3}{c}{\url{https://github.com/ratschlab/GP-VAE}.}\\
 & {\textbf{TSDE}} & \textbf{0.365(0.001)} & \textbf{0.331(0.001)} & \textbf{0.597(0.002)} & \textbf{0.403(0.001)} & \textbf{0.371(0.001)} & \textbf{0.657(0.001)} & \textbf{0.517(0.001)} & \textbf{0.476(0.001)} & \textbf{0.775(0.001)} & \multicolumn{3}{c}{We report the mean and standard } \\
 & {\textbf{TSDE+ft}} & \underline{0.374(0.001)} & \underline{0.338(0.001)} & \underline{0.610(0.003)} & {0.421(0.001)} & \underline{0.385(0.001)} & \underline{0.677(0.003)} & {0.570(0.004)} & {0.522(0.006)} & \underline{0.821(0.006)} & \multicolumn{3}{c}{deviation of three runs.} \\

\cmidrule{0-10}
\end{tabular}}
\vspace{-2pt}
\end{table*}

\subsection{Imputation, Interpolation and Forecasting}

\subsubsection{Imputation} 
\label{sec:exp-imputation}
We carry out imputation experiments on PhysioNet\footnote{PhysioNet, a healthcare dataset with 4,000 records of 35 variables over 48 hours, is processed and hourly sampled as \cite{NEURIPS2019_0b105cf1,tashiro2021csdi}, leading to $\sim$80\% missing rate. For testing, we randomly mask 10\%, 50\%, and 90\% of observed values to create ground-truth scenarios.
On this dataset, we pretrain TSDE for 2,000 epochs, followed by a 200-epoch finetuning with an imputation mask.}~\cite{silva2012predicting}
and PM2.5\footnote{PM2.5, an air quality dataset, features hourly readings from 36 Beijing stations over 12 months with artificially generated missing patterns. 
Adapting \cite{tashiro2021csdi}, each series spans 36 consecutive timestamps. 
On this dataset, we pretrain for 1,500 epochs and finetune for 100 epochs using a history mask as detailed in Algorithm~\ref{algo:history_mask}.}~\cite{yi2016st-mvl}.
TSDE is benchmarked against several state-of-the-art TS imputation models. 
These include BRITS~\cite{NEURIPS2018_734e6bfc}, a deterministic method using bi-directional RNN for correlation capture; V-RIN~\cite{9370004}, employing variational-recurrent networks with feature and temporal correlations for uncertainty-based imputation; GP-VAE~\cite{pmlr-v108-fortuin20a}, integrating Gaussian Processes with VAEs; and CSDI~\cite{tashiro2021csdi}, the top-performing model among the diffusion-based TS imputation models.
The model performance is evaluated using continuous ranked probability score (CRPS) to assess the fit of predicted outcomes with original data distributions, and two deterministic metrics -- mean absolute error (MAE) and the root mean square error (RMSE). 
Deterministic metrics are calculated using the median across all samples, and CRPS value is reported as the normalized average score for all missing values distributions (approximated with 100 samples).

The imputation results, as detailed in the upper part of Table ~\ref{tab_imputation_interpolation_results}, highlight TSDE's superior performance over almost all metrics, outperforming all baselines. 
Notably, the pretraining-only variant (i.e.,~``TSDE'') excels on the PhysioNet dataset, underpinning its robustness and enhanced generalization capability, even without the need of any imputation-specific finetuning.
For the PM2.5 dataset, finetuning TSDE (i.e.,~``TSDE+ft'') yields improved outcomes, likely attributable to its capability to adapt to the dataset's structured missing value patterns.
Overall, TSDE's improvement in CRPS by 4.2\%-6.5\% over CSDI, a leading diffusion-based TS imputation model, signifies a notable advancement in the field.
For a qualitative illustration of imputation results, refer to Figure~\ref{fig:tsde_imputation}.

\vspace{-2pt}
\subsubsection{Interpolation}
For interpolation analysis, we utilized the same PhysioNet dataset~\cite{silva2012predicting}, adopting the processing methods from~\cite{tashiro2021csdi,shukla2021multitime,NEURIPS2019_42a6845a}. 
Ground truth scenarios were created by masking all values at randomly selected timestamps, sampled at rates of 10\%, 50\% and 90\%. 
TSDE is pretrained for 2,000 epochs, and then further finetuned using an interpolation-only mask for another 200 epochs.
In our benchmarking, TSDE is compared against three TS interpolation methods: 
(1) Latent ODE~\cite{NEURIPS2019_42a6845a}, an RNN-based model leveraging ODE (ordinary differential equation) for dynamic, continuous and irregular TS handling; 
(2) mTANs~\cite{shukla2021multitime}, utilizing time embeddings and attention mechanisms, noted for its strong performance in irregular TS interpolation; and 
(3) CSDI~\cite{tashiro2021csdi} which has also reported competitive result in interpolation tasks.

The results in the lower section of Table ~\ref{tab_imputation_interpolation_results} demonstrate TSDE's exceptional performance in interpolation, outperforming CSDI by 3.6\%-7.0\% in CRPS, 5.8\%-8.6\% in MAE, and 6.1\%-17.3\% in RMSE. 
These findings highlight TSDE's adeptness in managing irregular timestamp gaps, a likely factor behind the observation that finetuning does not enhance the pretraining-only TSDE's performance. 
Comparatively, while CSDI also operates on a similar diffusion model backbone, TSDE's edge lies in its unique embedding learning ability via IIF masking, adeptly capturing intricate TS characteristics and dynamics for improved results.
A qualitative illustration of interpolation results can be found in Figure~\ref{fig:tsde_interpolation}.

\vspace{-2pt}
\subsubsection{Forecasting}
\label{sec:exp-forecasting}
We conducted two sets of benchmarking experiments. The first was a benchmarking for probabilistic multivariate time series forecasting. We employ five real-world datasets: (1) \textit{Electricity}, tracking hourly consumption across 370 customers; (2) \textit{Solar}, detailing photovoltaic production at 137 Alabama stations; (3) \textit{Taxi}, recording half-hourly traffic from 1,214 New York locations; (4) \textit{Traffic}, covering hourly occupancy rates of 963 San Francisco car lanes; and (5) \textit{Wiki}, monitoring daily views of 2,000 Wikipedia pages. 
Adapting the practices from \cite{NEURIPS2019_0b105cf1,Nguyen_Quanz_2021,tashiro2021csdi}, each dataset is converted into a series of multivariate sequences, with $L_{1}$ historical timestamps followed by $L_{2}$ timestamps for forecasting. 
Training data apply a rolling window approach with a stride of 1, while validation and testing data employ a stride of $L_{2}$, ensuring distinct, non-overlapping series for evaluation. 
Specific $L_{1}$ and $L_{2}$ values are outlined in Table ~\ref{tab:forecast_datasets}.
For evaluation metrics, we use CRPS and MSE, supplemented by CRPS-Sum, as introduced in \cite{NEURIPS2019_0b105cf1}. CRPS-Sum is computed by summing across different features, capturing the joint impact of feature distributions.
As of benchmarking baselines, we include several {\it state-of-the-art probabilistic MTS forecasting models}: GP-copula~\cite{NEURIPS2019_0b105cf1}, TransMAF~\cite{rasul2021multivariate} and TLAE~\cite{Nguyen_Quanz_2021}.
Additionally, in the realm of {\it diffusion-based methods}, we include CSDI~\cite{tashiro2021csdi} and TimeGrad~\cite{rasul2021autoregressive}.

For the second benchmarking, which is a deterministic benchmarking including recent baselines in the time series library \cite{wu2023timesnet}, we conducted five experiments for each baseline following the same setting in \cite{wu2023timesnet} with history-prediction window lengths of \{8-8, 16-16, 32-32, 96-96, 96-192\} on the Electricity dataset. We report the averaged performance in terms of MAE and MSE. We compared TSDE with the following baselines: TimesNet~\cite{wu2023timesnet}, ETSformer~\cite{woo2023etsformer}, LightTS~\cite{zhang2022more}, DLinear~\cite{Zeng2022AreTE}, FEDformer~\cite{pmlr-v162-zhou22g},  Non-stationary Transformer~\cite{NEURIPS2022_4054556f}, Autoformer~\cite{NEURIPS2021_bcc0d400}, Pyraformer~\cite{liu2022pyraformer}, Informer~\cite{Zhou_Zhang_Peng_Zhang_Li_Xiong_Zhang_2021},  Reformer~\cite{kitaev2020reformer}, and PatchTST~\cite{nie2023a}.

\begin{table*}[htbp]
  \caption{Forecasting task results on Electricity following~\cite{wu2023timesnet} setting. We compare extensive competitive models under five different history-prediction lengths \{8-8, 16-16, 32-32, 96-96, 96-192\}. \emph{Avg} is averaged from all five history-prediction lengths results. See Table~\ref{tab:tslib_full_forecasting_results} for full results.}\label{tab:full_forecasting_results}
  \vskip 0.05in
  \centering
  \resizebox{1\textwidth}{!}{
  \begin{threeparttable}
  \begin{small}
  \renewcommand{\multirowsetup}{\centering}
  \setlength{\tabcolsep}{1pt}
  \begin{tabular}{c|cc|cc|cc|cc|cc|cc|cc|cc|cc|cc|cc|cc}
    \toprule
    \multicolumn{1}{c}{\multirow{3}{*}{Models}} &
    \multicolumn{2}{c}{\rotatebox{0}{\scalebox{0.8}{\textbf{TSDE}}}} & 
    \multicolumn{2}{c}{\rotatebox{0}{\scalebox{0.8}{TimesNet}}} &
    \multicolumn{2}{c}{\rotatebox{0}{\scalebox{0.8}{{ETSformer}}}} &
    \multicolumn{2}{c}{\rotatebox{0}{\scalebox{0.8}{LightTS$^\ast$}}} &
    \multicolumn{2}{c}{\rotatebox{0}{\scalebox{0.8}{DLinear$^\ast$}}} &
    \multicolumn{2}{c}{\rotatebox{0}{\scalebox{0.8}{FEDformer}}} & \multicolumn{2}{c}{\rotatebox{0}{\scalebox{0.8}{Stationary}}} & \multicolumn{2}{c}{\rotatebox{0}{\scalebox{0.8}{Autoformer}}} & \multicolumn{2}{c}{\rotatebox{0}{\scalebox{0.8}{Pyraformer}}} &  \multicolumn{2}{c}{\rotatebox{0}{\scalebox{0.8}{Informer}}} & \multicolumn{2}{c}{\rotatebox{0}{\scalebox{0.8}{Reformer}}}   & \multicolumn{2}{c}{\rotatebox{0}{\scalebox{0.8}{PatchTST}}} \\
    \multicolumn{1}{c}{} & \multicolumn{2}{c}{\scalebox{0.8}{(\textbf{Ours})}} & 
    \multicolumn{2}{c}{\scalebox{0.8}{\citeyearpar{wu2023timesnet}}} &
    \multicolumn{2}{c}{\scalebox{0.8}{\citeyearpar{woo2023etsformer}}} &
    \multicolumn{2}{c}{\scalebox{0.8}{\citeyearpar{zhang2022more}}} & \multicolumn{2}{c}{\scalebox{0.8}{\citeyearpar{Zeng2022AreTE}}} & \multicolumn{2}{c}{\scalebox{0.8}{\citeyearpar{pmlr-v162-zhou22g}}} & \multicolumn{2}{c}{\scalebox{0.8}{\citeyearpar{NEURIPS2022_4054556f}}} & \multicolumn{2}{c}{\scalebox{0.8}{\citeyearpar{NEURIPS2021_bcc0d400}}} &  \multicolumn{2}{c}{\scalebox{0.8}{\citeyearpar{liu2022pyraformer}}} & 
    \multicolumn{2}{c}{\scalebox{0.8}{\citeyearpar{Zhou_Zhang_Peng_Zhang_Li_Xiong_Zhang_2021}}}& \multicolumn{2}{c}{\scalebox{0.8}{\citeyearpar{kitaev2020reformer}}}& \multicolumn{2}{c}{\scalebox{0.8}{\citeyearpar{nie2023a}}}\\
     \cmidrule(lr){2-3}\cmidrule(lr){4-5}\cmidrule(lr){6-7} \cmidrule(lr){8-9}\cmidrule(lr){10-11}\cmidrule(lr){12-13}\cmidrule(lr){14-15}\cmidrule(lr){16-17}\cmidrule(lr){18-19}\cmidrule(lr){20-21}\cmidrule(lr){22-23}\cmidrule(lr){24-25}
    \multicolumn{1}{c}{} & \scalebox{0.78}{MSE} & \scalebox{0.78}{MAE} & \scalebox{0.78}{MSE} & \scalebox{0.78}{MAE} & \scalebox{0.78}{MSE} & \scalebox{0.78}{MAE} & \scalebox{0.78}{MSE} & \scalebox{0.78}{MAE} & \scalebox{0.78}{MSE} & \scalebox{0.78}{MAE} & \scalebox{0.78}{MSE} & \scalebox{0.78}{MAE} & \scalebox{0.78}{MSE} & \scalebox{0.78}{MAE} & \scalebox{0.78}{MSE} & \scalebox{0.78}{MAE} & \scalebox{0.78}{MSE} & \scalebox{0.78}{MAE} & \scalebox{0.78}{MSE} & \scalebox{0.78}{MAE} & \scalebox{0.78}{MSE} & \scalebox{0.78}{MAE}  & \scalebox{0.78}{MSE} & \scalebox{0.78}{MAE} \\
    \toprule
    \scalebox{0.78}{Avg} &\scalebox{0.78}{\boldres{0.169}} &\scalebox{0.78}{\boldres{0.253
}} 
    &\scalebox{0.78}{0.195} &\scalebox{0.78}{0.290} &\scalebox{0.78}{0.241} &\scalebox{0.78}{0.355} &\scalebox{0.78}{0.227} &\scalebox{0.78}{0.329} &\scalebox{0.78}{0.359} &\scalebox{0.78}{0.408}
    &\scalebox{0.78}{0.193} &\scalebox{0.78}{0.310} &\scalebox{0.78}{0.191} &\scalebox{0.78}{0.285} &\scalebox{0.78}{0.192} &\scalebox{0.78}{0.311} &\scalebox{0.78}{0.299} &\scalebox{0.78}{0.367} &\scalebox{0.78}{0.295} &\scalebox{0.78}{0.390} &\scalebox{0.78}{0.287} &\scalebox{0.78}{0.382} 
    &\scalebox{0.78}{0.342}&\scalebox{0.78}{0.362}\\
    \bottomrule
  \end{tabular}
    \begin{tablenotes}
        \footnotesize
        \item[] $\ast$ means that there are some mismatches between our input-output setting and their papers. We adopt their official codes and only change the length of input and output sequences for a fair comparison.
  \end{tablenotes}
    \end{small}
  \end{threeparttable}
}
\end{table*}

\begin{table}[h!]
\footnotesize
\addtolength{\tabcolsep}{-4.5pt}
\caption[Results]{\small Probabilistic MTS forecasting results embodying both TSDE (pretraining-only) and finetuned (TSDE+ft) variants. Baseline results are either sourced or reproduced from \cite{NEURIPS2019_0b105cf1,rasul2021multivariate,Nguyen_Quanz_2021}. For TSDE-related experiments, we report the mean and SD across three iterations.}
\vspace{-6pt}
\label{tab:forecasting_results}
\centering
\scalebox{0.94}{
\begin{tabular}{l | l | c *{5}{S[table-format=2.3]}}
 \multirow{3}{*}{  } & \multirow{1}{*}{Models} & \multicolumn{1}{c}{Electricity} & \multicolumn{1}{c}{Solar} & \multicolumn{1}{c}{Taxi} & \multicolumn{1}{c}{Traffic} & \multicolumn{1}{c}{Wiki} \\
\midrule
\multirow{6}{*}{\rotatebox{90}{CRPS}} & {GP-copula} & {0.056(0.002)} & {0.371(0.022)} & {0.360(0.201)} & {0.133(0.001)} & {0.236(0.000)} \\
 & {TransMAF} & {0.052(0.000)} & \underline{0.368(0.001)} & {0.377(0.002)} & {0.134(0.001)} & {0.274(0.007)}  \\
 & {TLAE} & {0.058(0.003)} & \textbf{0.335(0.044)} & {0.369(0.011)} & {0.097(0.002)} & {0.298(0.002)}  \\
 & {CSDI} & {0.043(0.001)*} & {0.396(0.021)*}\textsuperscript{$\dagger$} & \underline{0.277(0.006)*} & \textbf{0.076(0.000)*} & {0.232(0.006)*}  \\
 & {\textbf{TSDE} } & \underline{0.043(0.000)} & {0.400(0.025)\textsuperscript{$\dagger$}} & \textbf{0.277(0.001)} & {0.091(0.001)} & \textbf{0.222(0.003)}  \\
 & {\textbf{TSDE+ft} } & \textbf{0.042(0.000)} & {0.375(0.013)\textsuperscript{$\dagger$}} & {0.282(0.001)} & \underline{0.081(0.001)} & \underline{0.226(0.003)}  \\
 \midrule
 \multirow{7}{*}{\rotatebox{90}{CRPS-sum}} & {GP-copula} & {0.024(0.002)} & {0.337(0.024)} & {0.208(0.183)} & {0.078(0.002)} & {0.086(0.004)} \\
& {TransMAF} & {0.021(0.000)} & {0.301(0.014)} & {0.179(0.002)} & {0.056(0.001)} & {0.063(0.003)}  \\
& {TimeGrad} & {0.021(0.001)} & \underline{0.287(0.020)} & \textbf{0.114(0.020)} & {0.044(0.006)} & \textbf{0.049(0.002)}  \\
& {TLAE} & {0.040(0.003)} & \textbf{0.124(0.057)} & \underline{0.130(0.010)} & {0.069(0.002)} & {0.241(0.001)}  \\
 & {CSDI} & \underline{0.019(0.001)*} & {0.345(0.029)*\textsuperscript{$\dagger$}} & {0.138(0.008)*} & \textbf{0.020(0.000)*} & {0.084(0.013)*} \\
 & {\textbf{TSDE}} & {0.020(0.001)} & {0.453(0.026)\textsuperscript{$\dagger$}} & {0.136(0.003)} & {0.038(0.003)} & {0.064(0.002)} \\
  & {\textbf{TSDE+ft}} & \textbf{0.017(0.001)} & {0.345(0.012)\textsuperscript{$\dagger$}} & {0.153(0.006)} & \underline{0.025(0.001)} & \underline{0.059(0.003)}  \\

\midrule
\multirow{6}{*}{\rotatebox{90}{MSE}} & {GP-copula} & {2.4e5(5.5e4)} & {9.8e2(5.2e1)} & {3.1e1(1.4e0)} & {6.9e-4(2.2e-5)} & {4.0e7(1.6e9)} \\
& {TransMAF} & {2.0e5} & {9.3e2} & {4.5e1} & {5.0e-4} & \textbf{3.1e7}  \\
& {TLAE} & {2.0e5(1.6e4)} & \textbf{6.8e2(1.3e2)} & {2.6e1(1.4e0)} & {4.0e-4(5.0e-6)} & {3.8e7(7.2e4)}  \\
 & {CSDI} & {1.23e5(9.7e3)*} & {1.12e3(1.2e2)*}\textsuperscript{$\dagger$} & \textbf{1.82e1(7.8e-1)*} & \textbf{3.64e-4(0.0e0)*} & {4.43e7(1.0e7)*}  \\
 & {\textbf{TSDE}} & \underline{1.20e5(3.5e3)} & {1.07e3(9.8e1)\textsuperscript{$\dagger$}} & \underline{1.89e1(3.7e-1)} & {4.34e-4(0.0e0)} & \underline{3.59e7(7.2e4)}  \\
 & {\textbf{TSDE+ft}} & \textbf{1.16e5(6.0e3)} & \underline{9.25e2(4.9e1)\textsuperscript{$\dagger$}} & {1.92e1(2.4e-1)} & \underline{3.88e-4(0.0e0)} & {3.62e7(1.8e5)}  \\
\midrule
\end{tabular}}
\begin{flushleft}
\vspace{-0.12cm}
\scriptsize
\ * We replace the linear Transformers \cite{linearattentiontrans} in CSDI with the Pytorch TransformerEncoder~\cite{NEURIPS2019_bdbca288}.

\ $\dagger$ We take the training MTS dataset and split it into training, validation and testing sets.
\end{flushleft}
\vspace{-4pt}
\end{table}

The forecasting results, as detailed in Table~\ref{tab:forecasting_results}, showcase TSDE's robust performance, especially when finetuned with a forecasting mask.
Its effectiveness is notable when compared to CSDI, which is the most closely related method, sharing a diffusion backbone.
TSDE particularly excels in the Electricity, Taxi, and Wiki datasets, especially as evaluated by the CRPS metric.
However, it is important to note a discrepancy in the Solar dataset performance between TSDE/CSDI and other baselines, likely due to a data split issue: the actual test set, per the source code, is identical to the training set, which contradicts the details reported in the corresponding paper. Table~\ref{tab:full_forecasting_results} demonstrates that TSDE outperforms the recent baselines in terms of average MSE and MAE, highlighting its robustness and superiority compared to recent methods. The detailed results for each window length are available in the appendix, Table~\ref{tab:tslib_full_forecasting_results}. 
For a qualitative illustration, refer to Figure~\ref{fig:tsde_forecasting}.

\vspace{-2pt}
\subsubsection{Ablation Study}
In an ablation study on TSDE across imputation, interpolation, and forecasting, evaluated on PhysioNet (10\% missing ratio) and Electricity datasets, two configurations were tested: one without crossover, and another without IIF mask (replaced by an imputation mask detailed in Algorithm~\ref{algo:imputation_masking}). Table ~\ref{tab:ablations_results} underscores the positive contribution of the crossover mechanism across all three tasks. The impact of IIF masking, while less pronounced for imputation and interpolation, becomes noticeable in the forecasting task. This can be attributed to the random PhysioNet missing values, which are distributed fundamentally differently from a typical forecasting scenario. 
Thus, IIF strategy is important for TSDE to gain a generalization ability across various settings. 
The contrast between ``{\small TSDE}'' and ``{\small TSDE+ft}'' in Tables~\ref{tab_imputation_interpolation_results} and \ref{tab:forecasting_results} serves as an ablation study for finetuning; it reveals that pretrained TSDE can achieve competitive results without the necessity of finetuning.

\begin{table}[h!]
\footnotesize
\addtolength{\tabcolsep}{-2.5pt}
\caption[Results]{\small Ablation study on PhysioNet (imputation and interpolation) and Electricity (forecasting) datasets.} 
\vspace{-6pt}
\label{tab:ablations_results}
\centering

\scalebox{0.9}{\begin{tabular}{ l | c *{3}{S[table-format=2.3]}}

  {\shortstack{Ablation\\Configuration}} & \multicolumn{1}{c}{\shortstack{Imputation\\(MAE/CRPS)}} & \multicolumn{1}{c}{\shortstack{Interpolation\\(MAE/CRPS)}} & \multicolumn{1}{c}{\shortstack{Forecasting\\(CRPS-sum/CRPS)}}\\
\midrule
{w/o crossover} & {0.252(0.001)/0.274(0.001)} & {0.339(0.000)/0.373(0.000)} & {0.021(0.001)/0.046(0.001)} \\
{w/o IIF mask} & {0.207(0.001)/0.225(0.001)} & {0.330(0.001)/0.364(0.001)} & {0.028(0.004)/0.053(0.003)}\\
{TSDE} & {0.208(0.001)/0.226(0.002)} & {0.331(0.001)/0.365(0.001)}&  {{\bf0.020}(0.001)/{\bf0.043}(0.000)} \\

\midrule
\end{tabular}}
\vspace{-8pt}
\end{table}
\vspace{-2pt}
\subsubsection{Inference Efficiency}
\label{sec:exp-inf-efficiency}
\begin{figure*}[h]
    \begin{minipage}{0.65\linewidth}
        \subfigure[Imputation on PhysioNet\label{fig:tsde_imputation}]{\includegraphics[height=2.6cm]{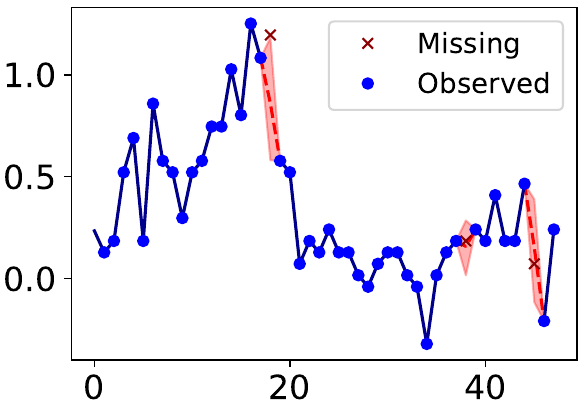}}
        \subfigure[Interpolation on Electricity\label{fig:tsde_interpolation}]{\includegraphics[height=2.6cm]{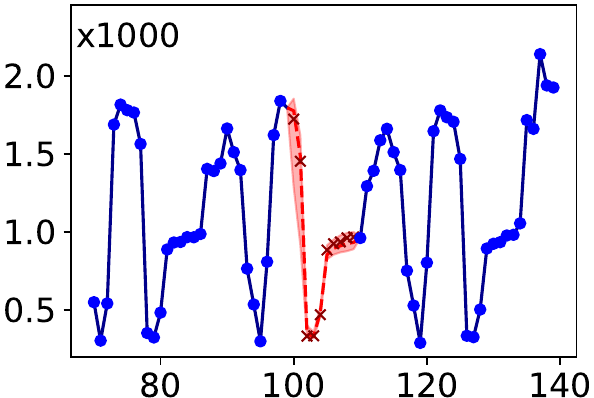}}
        \subfigure[Forecasting on Electricity\label{fig:tsde_forecasting}]{\includegraphics[height=2.6cm]{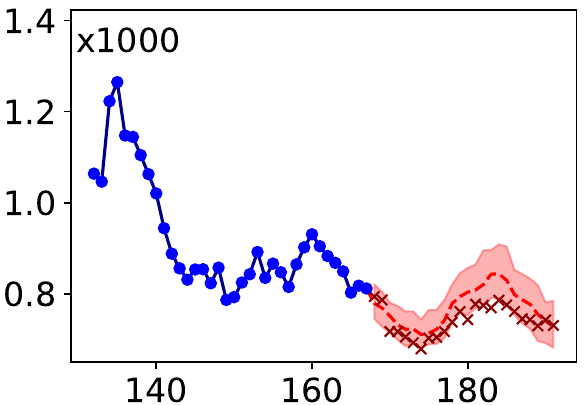}}
        \vspace{-15pt}
        \caption{\small Comparison of predicted and ground truth values for (a) imputation (10\% missing), (b) interpolation, and (c) forecasting. The line is the median of the predictions and the red shade indicates 5\%$\sim$95\% quantile for missing/future values. See Appendix~\ref{appendix-vis} for more results.}\label{fig:tsde_pred_viz}
    \end{minipage}
    \hspace{15pt}
    \begin{minipage}{0.3\linewidth}
        \centering
        \begin{table}[H]
        \begin{tabular}{ l | c *{2}{S[table-format=2.3]}}
        
          \multirow{1}{*}{Datasets} & \multicolumn{1}{c}{CSDI* (sec.)} & \multicolumn{1}{c}{TSDE (sec.)}\\
        \midrule
        
         {Electricity} & {1,997} & {163}\\
         {Solar} & {608} & {62}\\
         {Taxi} & {27,533} & {1,730}\\
         {Traffic} & {7,569} & {422}\\
         {Wiki} & {9,138} & {391}\\
        \midrule
        \end{tabular}
        \begin{flushleft}
        \scriptsize
        \ * For fair comparison, the linear Transformer encoders in CSDI~\cite{tashiro2021csdi} is replaced with the TransformerEncoder~\cite{NEURIPS2019_bdbca288} implementation in Pytorch.
        \end{flushleft}
        \vspace{3pt}
        \caption{\small Inference time comparison for forecasting tasks between TSDE and CSDI.}\label{tab:inference_time}
        \end{table}
    \end{minipage}
    \vspace{-5pt}
\end{figure*}

Similar to CSDI~\cite{tashiro2021csdi}, TSDE performs inference by gradual denoising from the last diffusion step $T$=50 to the initial step $t$=1, to approximate the true data distribution of missing or future values for imputation/interpolation/forecasting tasks. 
Typically, this iterative process can become computationally expensive. 
TSDE achieves a substantial acceleration in this process as illustrated in Table~\ref{tab:inference_time}, where TSDE is ten times faster than CSDI under the same experimental setup. This is primarily owing to its globally shared, efficient dual-orthogonal Transformer encoders with a crossover mechanism, merely requiring approximately a quarter of the parameters used by CSDI for MTS encoding.

\begin{table*}[!h]
    \begin{minipage}{0.68\linewidth}
        \caption[Results]{\small Anomaly detection: baseline results are cited from Table 27 of \cite{zhou2023one}; higher scores indicate better performance; the best and second best results are in bold and underlined, respectively.}
\vspace{-7pt}
\label{tab:ping_ad_results}
\centering
\small
\addtolength{\tabcolsep}{-3.4pt}
\renewcommand{\arraystretch}{0.8}
    \begin{tabular}{l | c c c | c c c | c c c | c c c | c c c | c }
        \multirow{2}{*}{Models} & & {SMD} & & &  {MSL} & & & {SMAP} & & & {SWaT} & & & {PSM} & &  \multirow{2}{*}{\shortstack{Avg.\\ F1}}\\
         
         & {P}  & {R} & {F1} & {P}  & {R} & {F1} & {P}  & {R} & {F1} & {P}  & {R} & {F1} & {P}  & {R} & {F1} & \\  
        \midrule
        {Transformer} & {83.6} & {76.1} & {79.6} & {71.6} & \underline{87.4} & {78.7} & {89.4} & {57.1} & {69.7} & {68.8} & {96.5} & {80.4} & {62.7} & {96.6} & {76.1} & \textcolor{blue}{76.9} \\ 
        {LogSparseT.} & {83.5} & {70.1} & {76.2} & {73.0} & \underline{87.4} & {79.6} & {89.1} & {57.6} & {70.0} & {68.7} & \textbf{97.3} & {80.5} & {63.1} & \textbf{98.0} & {76.7} & \textcolor{blue}{76.6} \\ 
        {Reformer} & {82.6} & {69.2} & {75.3} & {85.5} & {83.3} & {84.4} & {90.9} & {57.4} & {70.4} & {72.5} & {96.5} & {82.8} & {59.9} & {95.4} & {73.6} & \textcolor{blue}{77.3} \\ 
        {Informer} & {86.6} & {77.2} & {81.6} & {81.8} & {86.5} & {84.1} & {90.1} & {57.1} & {69.9} & {70.3} & \underline{96.7} & {81.4} & {64.3} & {96.3} & {77.1} & \textcolor{blue}{78.8} \\ 
        {AnomalyT.\textsuperscript{$\dagger$}} & \textbf{88.9} & {82.2} & \underline{85.5} & {79.6} & \underline{87.4} & {83.3} & {91.8} & {58.1} & \underline{71.2} & {72.5} & \textbf{97.3} & {83.1} & {68.3} & {94.7} & {79.4} & \textcolor{blue}{80.5} \\ 
        {Pyraformer} & {85.6} & {80.6} & {83.0} & {83.8} & {85.9} & {84.9} & \underline{92.5} & {57.7} & {71.1} & {87.9} & {96.0} & {91.8} & {71.7} & {96.0} & {82.1} & \textcolor{blue}{82.6} \\ 
        {Autoformer} & {88.1} & {82.3} & {85.1} & {77.3} & {80.9} & {79.0} & {90.4} & {58.6} & {71.1} & {89.8} & {95.8} & {92.7} & \underline{99.1} & {88.1} & {93.3} & \textcolor{blue}{84.3} \\ 
        {NonStation.} & {88.3} & {81.2} & {84.6} & {68.5} & \textbf{89.1} & {77.5} & {89.4} & \underline{59.0} & {71.1} & {68.0} & \underline{96.7} & {79.9} & {97.8} & {96.8} & \underline{97.3} & \textcolor{blue}{82.1} \\ 
        {DLinear} & {83.6} & {71.5} & {77.1} & {84.3} & {85.4} & {84.9} & {92.3} & {55.4} & {69.3} & {80.9} & {95.3} & {87.5} & {98.3} & {89.3} & {93.5} & \textcolor{blue}{82.5} \\ 
        {LightTS} & {87.1} & {78.4} & {82.5} & {82.4} & {75.8} & {78.9} & \textbf{92.6} & {55.3} & {69.2} & {92.0} & {94.7} & \underline{93.3} & {98.4} & {96.0} & {97.1} & \textcolor{blue}{84.2} \\ 
        {FEDformer} & {87.9} & \underline{82.4} & {85.1} & {77.1} & {80.1} & {78.6} & {90.5} & {58.1} & {70.8} & {90.2} & {96.4} & {93.2} & {97.3} & \underline{97.2} & {97.2} & \textcolor{blue}{85.0} \\ 
        {ETSformer} & {87.4} & {79.2} & {83.1} & {85.1} & {84.9} & \underline{85.0} & {92.2} & {55.7} & {69.5} & {90.0} & {80.4} & {84.9} & \textbf{99.3} & {85.3} & {91.8} & \textcolor{blue}{82.9} \\ 
        {PatchTS.} & {87.3} & {82.1} & {84.6} & {88.3} & {71.0} & {78.7} & {90.6} & {55.5} & {68.8} & {91.1} & {80.9} & {85.7} & {98.8} & {93.5} & {96.1} & \textcolor{blue}{82.8} \\ 
        {TimesNet*} & {87.9} & {81.5} & {84.6} & \underline{89.5} & {75.4} & {81.8} & {90.1} & {56.4} & {69.4} & {90.7} & {95.4} & {93.0} & {98.5} & {96.2} & \textbf{97.3} & \textcolor{blue}{85.2} \\
        {GPT4TS\textsuperscript{$\ddagger$}} & \underline{88.9} & \textbf{85.0} & \textbf{86.9} & {82.0} & {82.9} & {82.4} & {90.6} & \textbf{60.9} & \textbf{72.9} & \underline{92.2} & {96.3} & \textbf{94.2} & {98.6} & {95.7} & {97.1} & \textbf{\textcolor{blue}{86.7}} \\
        {\textbf{TSDE}\textsuperscript{$\ddagger$}} & {87.5} & {82.2} & {84.8} & \textbf{90.1} & {84.5} & \textbf{87.2} & {91.4} & {56.9} & {70.1} & \textbf{98.2} & {92.9} & {92.5} & {98.6} & {90.7} & {94.5} & \underline{\textcolor{blue}{85.8}} \\ 
        \midrule
    \end{tabular}
\begin{flushleft}
\vspace{-0.12cm}
\scriptsize
\ * Reproduced with \url{https://github.com/thuml/Time-Series-Library}.
\ $\dagger$ Reconstruction error is used as joint criterion for fair comparison.

\ $\ddagger$ GPT4TS leverage a pretrained LLM (GPT-2) with 1.5B parameters, while TSDE merely uses two single-layer Transformer encoders.
\end{flushleft}
    \end{minipage}
    \hspace{15pt}
    \begin{minipage}{0.27\linewidth}
        \caption[Results]{\small Classification performance on PhysioNet measured with AUROC. The baseline results are sourced from Table~2 of \cite{NEURIPS2018_734e6bfc} and Table~3 of \cite{pmlr-v108-fortuin20a}.}
\label{tab:classification_results}
\vspace{-10pt}
\centering
\addtolength{\tabcolsep}{1pt}
\renewcommand{\arraystretch}{0.94}
\small
\begin{tabular}{ l c *{1}{S[table-format=2.3]}}
  \multirow{1}{*}{Models} & \multicolumn{1}{c}{AUROC} \\
\midrule
{Mean imp.~\cite{pmlr-v108-fortuin20a,10.5555/21412}} & {0.70 $\pm$ 0.000 }  \\
{Forward imp.~\cite{pmlr-v108-fortuin20a,10.5555/21412}} & {0.71 $\pm$ 0.000}  \\
{GP~\cite{Rasmussen2004}} & {0.70 $\pm$ 0.007}  \\
{VAE~\cite{DBLP:journals/corr/KingmaW13,10.5555/21412}} & {0.68 $\pm$ 0.002}  \\
{HI-VAE~\cite{nazabal2020handling}} & {0.69 $\pm$ 0.010}  \\
{GRUI-GAN~\cite{NEURIPS2018_96b9bff0}} & {0.70 $\pm$ 0.009}  \\
{GP-VAE~\cite{pmlr-v108-fortuin20a}} & {0.73 $\pm$ 0.006}  \\
{GRU-D~\cite{Che2018}} & \underline{0.83 $\pm$ 0.002}  \\
{M-RNN~\cite{Yoon2017MultidirectionalRN}} & {0.82 $\pm$ 0.003}  \\
{BRITS-LR~\cite{NEURIPS2018_734e6bfc}}\textsuperscript{$\dagger$} & {0.74 $\pm$ 0.008}  \\
{BRITS-RF~\cite{NEURIPS2018_734e6bfc}}* & {0.81 $\pm$ (N/A)} \\
{BRITS~\cite{NEURIPS2018_734e6bfc}} & \textbf{0.85 $\pm$ 0.002}  \\
{\textbf{TSDE}} & \textbf{0.85 $\pm$ 0.001}  \\
\midrule
\end{tabular}
\begin{flushleft}
\vspace{-0.12cm}
\scriptsize
\ $\dagger$ Logistic Regression (LR) on imputed PhysioNet data.

\ * Train Random Forest (RF) on imputed PhysioNet data.
\end{flushleft}
\end{minipage}
\end{table*}

\vspace{-2pt}
\subsection{Anomaly Detection}
\label{sec:exp-anomaly-detection}
For anomaly detection, we adopt an unsupervised approach using reconstruction error as the anomaly criterion, aligning with \cite{wu2023timesnet, zhou2023one}. We evaluate TSDE on five benchmark datasets: SMD~\cite{10.1145/3292500.3330672}, MSL~\cite{10.1145/3219819.3219845}, SMAP~\cite{10.1145/3219819.3219845}, SWaT~\cite{7469060} and PSM~\cite{10.1145/3447548.3467174}.
Once TSDE is pretrained, a projection layer, designed to reconstruct MTS from TSDE embeddings, is finetuned by minimizing MSE reconstruction loss.
Our anomaly detection experiments align with TimesNet~\cite{zhou2023one}, utilizing preprocessed datasets from \cite{xu2022anomaly}. 
Following their method, we segment datasets into non-overlapping MTS instances of 100 timestamps each, labeling timestamps as anomalous based on a MSE threshold. 
This threshold is set according to the anomaly proportion in the validation dataset, ensuring consistency with baseline anomaly ratios for a fair comparison.

In this task, TSDE is benchmarked against an extensive set of baselines featuring diverse backbones, including a) {\it Frozen pretrained LLM-based models}: GPT4TS~\cite{zhou2023one}; 
b) {\it Task-agnostic foundation models}: TimesNet~\cite{wu2023timesnet}; 
c) {\it MLP (multi-layer perceptron) based models}: LightTS~\cite{zhang2022more} and DLinear~\cite{Zeng2022AreTE}; 
and finally d) {\it Transformer-based models}: Transformer~\cite{NIPS2017_3f5ee243}, Reformer~\cite{kitaev2020reformer}, Informer~\cite{Zhou_Zhang_Peng_Zhang_Li_Xiong_Zhang_2021}, Autoformer~\cite{NEURIPS2021_bcc0d400}, Pyraformer~\cite{liu2022pyraformer}, LogSparse Transformer~\cite{10.5555/3454287.3454758}, FEDformer~\cite{pmlr-v162-zhou22g}, Non-stationary Transformer~\cite{NEURIPS2022_4054556f}, ETSformer~\cite{woo2023etsformer}, PatchTST~\cite{nie2023a} and Anomaly Transformer~\cite{xu2022anomaly}.
The results in Table ~\ref{tab:ping_ad_results} reveal that TSDE's anomaly detection performance surpasses nearly all baselines, with less than a 1\% F1 score difference from GPT4TS. 
Notably, while TSDE doesn't outperform GPT4TS, it's important to consider that GPT4TS benefits from a pretrained LLM (GPT-2) with about 1.5 billion model parameters. 
TSDE, in contrast, relies on just two single-layer Transformer encoders (<0.3 million parameters), demonstrating its competitive edge despite having significantly fewer model parameters.

\subsection{Classification}
\label{sec:exp-classification}
To further inspect the discriminative power of the pretrained TSDE embedding, we utilize the labeled PhysioNet dataset to evaluate TSDE's performance on a binary classification downstream task.
This dataset, marked by in-hospital mortality labels for each patient, features MTS with over 80\% missing values. 
To address this, we pretrain TSDE for 2,000 epochs to impute the raw MTS. 
Subsequently, we train a simple MLP for 40 epochs to perform mortality classification. 
Given the imbalanced nature of PhysioNet labels, we assess our model's efficacy with AUROC as in \cite{NEURIPS2018_734e6bfc,pmlr-v108-fortuin20a}.
We benchmark TSDE against a diverse range of established MTS classification methods, categorized into 3 groups with a total of 12 methods: 
(1) heuristic methods: mean/forward imputation~\cite{pmlr-v108-fortuin20a,10.5555/21412}, (2) GP/VAE based models: GP~\cite{Rasmussen2004}, VAE~\cite{DBLP:journals/corr/KingmaW13}, HI-VAE~\cite{nazabal2020handling}, GP-VAE~\cite{pmlr-v108-fortuin20a}, and (3) RNN based models: GRUI-GAN~\cite{NEURIPS2018_96b9bff0}, GRU-D~\cite{Che2018}, M-RNN~\cite{Yoon2017MultidirectionalRN} and BRITS variants \cite{NEURIPS2018_734e6bfc}.

As shown in Table ~\ref{tab:classification_results}, TSDE surpasses all existing baselines and is on par with the state-of-the-art BRITS baseline. 
It is worth noting that BRITS achieves that performance by employing a sophisticated multi-task learning mechanism tailored for classification tasks. 
In contrast, our method achieves top-tier results by simply finetuning a simple MLP.
TSDE's remarkable performance, especially in challenging classification scenarios with significant class imbalance ($\sim$10\% positive classes), highlights its ability to learn generic embeddings well-suited for downstream MTS classification tasks.

\vspace{-4pt}
\subsection{Clustering}
\label{sec:exp-clustering}
MTS data often lack annotations, making supervised learning inapplicable.
In such scenarios, unsupervised clustering is a valuable method for uncovering intrinsic patterns and classes. 
We utilize the same pretrained TSDE model from our classification experiments (trained on PhysioNet with a 10\% missing ratio) to evaluate the clustering performance of TSDE embeddings.
Initially, we generate MTS embeddings using TSDE's pretrained embedding function.
For simplicity and visual clarity, these embeddings are projected into a 2D space using UMAP (uniform manifold approximation and projection)~\cite{McInnes2018}.
Subsequently, DBSCAN (density-based spatial clustering of applications with noise)~\cite{ester1996density} is applied to these 2D projections to obtain clusters.\\
\begin{figure}[h!]
\centering
    \subfigure[Raw MTS\label{fig:cluster-raw}]{\put(1,81){\color{black} \footnotesize ARI=0.010 \; \; RI=0.502}\includegraphics[height=2.75cm]{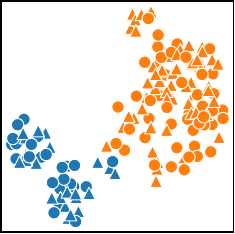}} 
\subfigure[Embed raw MTS\label{fig:cluster-raw-embed}]{\put(0,81){\color{black} \footnotesize ARI=\textbf{0.426} \; \; RI=\textbf{0.737}}\includegraphics[height=2.75cm]{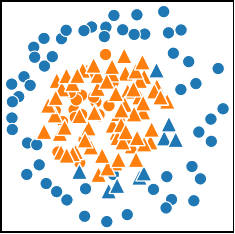}}
\subfigure[Embed imputed MTS\label{fig:cluster-imputed-embed}]{\put(1,81){\color{black} \footnotesize ARI=\underline{0.302} \; \; RI=\underline{0.684}}\includegraphics[height=2.75cm]{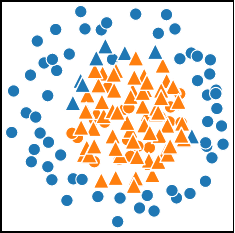}}
\vspace{-14pt}
\caption{\small Clustering of (a) raw MTS, (b) TSDE embedding of raw MTS, and (c) TSDE embedding of TSDE-imputed MTS. Marker shapes denote ground-truth binary labels; colors indicate DBSCAN~\cite{ester1996density} clusters after UMAP~\cite{McInnes2018} dimension reduction.} \label{fig:cluster-viz}
\vspace{-4pt}
\end{figure}

As shown in Figure~\ref{fig:cluster-viz}, the clustering quality is assessed across three data types: (a) raw MTS, (b) TSDE embeddings of raw MTS, and (c) TSDE embeddings of TSDE-imputed MTS.
The ground truth binary labels are indicated using two distinct marker shapes: circles and triangles.
When using raw MTS as seen in \ref{fig:cluster-raw}, the clusters are unfavourably intertwined, with data points from both classes intermingling.
However, the TSDE embeddings, whether derived from raw or imputed MTS, exhibit substantially improved cluster differentiation.
These embeddings enable more precise alignment with ground truth classifications, implying the capability of TSDE in capturing data nuances.
Furthermore, the negligible performance disparity between Figures~\ref{fig:cluster-raw-embed} and \ref{fig:cluster-imputed-embed} suggests that TSDE embeddings can be directly used for MTS clustering without the need of imputation.
This consistency is likely because our encoders proficiently encapsulate missing data traits, seamlessly integrating these subtleties into the embeddings.
To provide a quantitative assessment of clustering, given the presence of labels, we calculate RI (rand index)~\cite{6fd45ff5-8145-3f39-a258-8e3ef378c6a4} and ARI (adjusted RI)~\cite{Hubert1985}.
These metrics are reported on top of each setup in Figure~\ref{fig:cluster-viz}. Notably, the RI and ARI values align with the qualitative observations discussed earlier, further substantiating our findings.

\subsection{Embedding Visualization}

\begin{figure}[h!]
\centering
\includegraphics[width=0.47\textwidth]{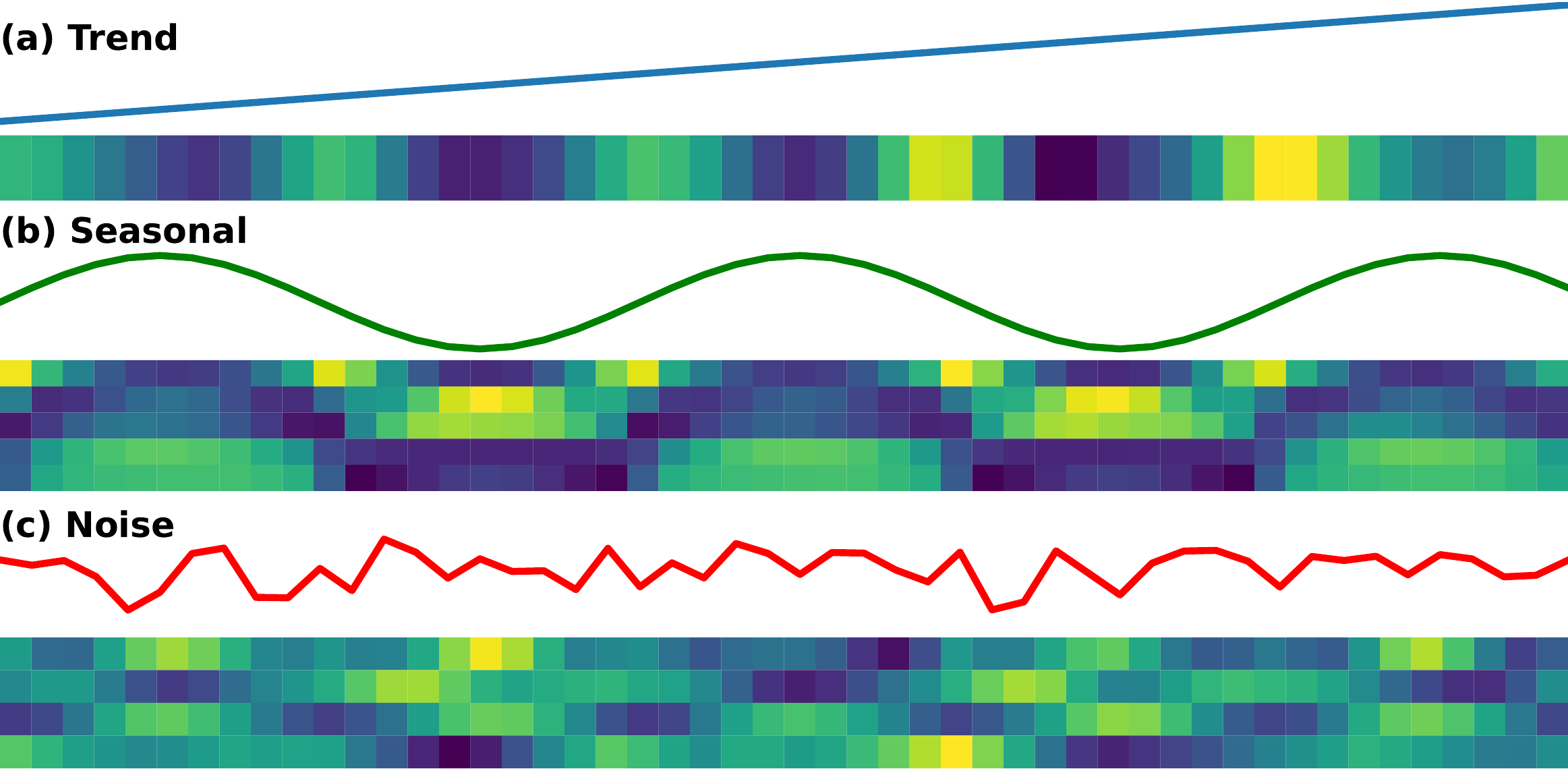}
\vspace{-8pt}
\caption{\small TSDE embedding visualization of (a) Trend, (b) Seasonal, and (c) Noise components from synthetic MTS.} \label{fig:embed-viz}
\end{figure}

To substantiate the representational efficacy of TSDE embeddings, we undertake a visualization experiment on synthetic MTS data, as showcased in Figure~\ref{fig:embed-viz}.
The data comprises three distinct UTS: (a) a consistently ascending trend, (b) a cyclical seasonal signal, and (c) a white noise component.
Each UTS embedding has two dimensions ($L\times 33$);
for a lucid depiction, we cluster the second dimension by treating it as 33 samples each of length $L$, and visualize the centroid of these clusters.
Intriguingly, the embeddings, which were pretrained on the entire synthetic MTS, vividly encapsulate the joint encoding effects of all series.
The trend's embedding delineates the series' progression, evident from the gradual color saturation changes, embodying the steady evolution.
The seasonal signal's embedding mirrors its inherent cyclicity, with color oscillations reflecting its periodic nature. 
Finally, the noise component's embeddings exhibit sporadic color band patterns (with subtle traces of seasonal patterns), capturing the inherent randomness. 

\section{Conclusion}
\label{sec:conclusion}
In this paper, we propose TSDE, a novel SSL framework for TSRL. 
TSDE, the first of its kind, effectively harnesses a diffusion process, conditioned on an innovative dual-orthogonal Transformer encoder architecture with a crossover mechanism, and employs a unique IIF mask strategy. 
Our comprehensive experiments across diverse TS analysis tasks, including imputation, interpolation, forecasting, anomaly detection, classification, and clustering, demonstrate TSDE's superior performance compared to state-of-the-art models. 
Specifically, TSDE shows remarkable results in handling MTS data with high missing rates and various complexities, thus validating its effectiveness in capturing the intricate MTS dynamics. 
Moreover, TSDE not only significantly accelerates inference speed but also showcases its versatile embeddings through qualitative visualizations, encapsulating key MTS characteristics. 
This positions TSDE as a robust, efficient, and versatile advancement in MTS representation learning, suitable for a wide range of MTS tasks.
Future work will focus on several key directions to address the limitation of slower inference for IIF tasks. Particularly, we will explore simplifying TSDE's architecture with a simple MLP without the need for the diffusion block, enabling the pretrained TSDE to execute IIF tasks independently. 

\begin{acks}
    We would like to thank Yang Song (OpenAI \& Stanford University) for his help to connect us with CSDI authors and the insightful discussions regarding time series representation learning.

    R. Tu would like to acknowledge the support of Gustav Eje Henter, Hedvig Kjellström and the Wallenberg AI, Autonomous Systems and Software Program (WASP). R. Tu was also funded by the Industrial Strategic Technology Development Program (grant no. 20023495) from MOTIE, Korea.
\end{acks}
\newpage
\bibliographystyle{ACM-Reference-Format}
\balance
\bibliography{ref}

\newpage
\appendix
\label{appendix_full}

\section{Datasets}
\label{appendix}

\subsection{Datasets for Imputation and Interpolation}
Overall specifications of the datasets used for imputation and interpolation are provided in Table~\ref{tab:data_imputation_interpolation}.
\begin{table}[h!]
\caption[imputation]{\small Overall specification of datasets used for imputation and interpolation tasks. ``pre'': pretraining; ``ft'': finetuning.} 
\label{tab:data_imputation_interpolation}
\centering
\footnotesize
\begin{tabular}{l | c c c c c c c c}
Dataset & \multicolumn{1}{c}{$K$} & \multicolumn{1}{c}{$L$} & \multicolumn{1}{c}{$N$} & \multicolumn{1}{c}{$N_{\text{train}}$} & \multicolumn{1}{c}{$N_{\text{val}}$} & \multicolumn{1}{c}{$N_{\text{test}}$} & \multicolumn{1}{c}{$E_{\text{pre}}$} & \multicolumn{1}{c}{$E_{\text{ft}}$} \\
\midrule
PhysioNet & 35 & 48 & 3,997 & 2,799 & 399 & 799 & 2,000 & 200 \\
PM2.5 & 36 & 36 & {-} & 4,842 & 709 & 82 & 1,500 & 100\\
\midrule
\end{tabular}
\vspace{-4pt}
\end{table}
\subsubsection{PhysioNet}
The PhysioNet dataset, part of the PhysioNet Challenge 2012 \cite{silva2012predicting}, is a rich repository of clinical time series data derived from intensive care unit (ICU) patients. This comprehensive dataset encompasses a total of 12,000 patients records, each comprising 42 recorded vital variables over 48 hours with an in-hospital death label indicating the survival of the patient. Patients in this dataset are categorized into three groups: training set A, open test set B, and hidden test set C, each containing 4,000 patients.
Aligning with the baselines established in previous works and evaluating on the PhysioNet benchmark \cite{NEURIPS2018_734e6bfc, NEURIPS2019_0b105cf1, tashiro2021csdi},  we use specifically ``set A'' of the dataset, and covert it to an hourly granularity, culminating in a format of 48 time steps for each series.
Furthermore, we select a subset of variables (vital signals) from this dataset, 35 out of 42, to facilitate a rigorous and meaningful benchmarking of our proposed model against the backdrop of existing research baselines. 

A notable characteristic of the PhysioNet dataset is its sparsity and significant proportion of missing values, $\sim$80\%, posing a unique challenge for data imputation and analysis techniques. Moreover, the classification labels are unbalanced with only around 10\% positive class labels. To prepare the data for our model, we adopt a systematic segmentation of the available data samples by splitting them into 5 folds. PhysioNet ``set A'' is systematically segmented into 20\% test, 10\% validation, and 70\% training sets, similar to \cite{tashiro2021csdi, NEURIPS2018_734e6bfc}.
As the vital signs (features) have different scales, we use the normalization process in  \cite{NEURIPS2018_734e6bfc}, resulting in features with zero mean and unit variance.  
To simulate ground-truth testing data for imputation task, we randomly mask a percentage (10\%, 50\% and 90\%) of the observed values in each MTS.
Conversely, for the interpolation task, a percentage of timestamps is randomly chosen, and all values corresponding to these selected timestamps are masked.

\vspace{-2pt}
\subsubsection{Air Quality PM2.5}
The air quality dataset \cite{yi2016st-mvl} collected from 36 monitoring stations in Beijing, offers hourly PM2.5 readings from 2014/05/01 to 2015/04/30. With each station representing a distinct feature, there are 36 features in total, highlighting the dataset’s spatial correlation. By setting 36 consecutive time steps as one time series, the temporal correlation will be captured as well. 
In an added layer of complexity, more intentional missing patterns were introduced (13\% missing values), hinting a structured absence and correlations of the actual data \cite{yi2016st-mvl}. 
For the experiments, data from the 3rd, 6th, 9th, and 12th months of the calendar year will be used for testing as in \cite{tashiro2021csdi, NEURIPS2018_734e6bfc}. 
Data from the remaining months will be allocated for training, with one month designated as the validation set.
During training and validation, a sliding window of size 36 with a stride of 1 is used to generate sequences for each calendar month in the respective set. 
During testing, a sliding window of the same size is employed, but with the stride set to the sequence length. 
If the length of monthly data in the test set was not divisible by 36, we allow the final sequence to overlap with its predecessor and excluded the results for the overlapping sections from aggregation.

\vspace{-4pt}
\subsection{Forecasting Datasets}

The forecasting datasets exhibit diverse dimensionality, featuring a range of hundreds to thousands of features. 
Preprocessed by \citet{NEURIPS2019_0b105cf1}, these datasets are available as JSON files, pre-split into training and testing sets. Moreover, accompanying metadata details the JSON file contents and specifies the value frequency for each series. 
Each line corresponds to one variable, and the target values are the measurements of that variable. 
The specific processing details for each dataset can be found in Table~\ref{tab:forecast_datasets}. 
The objective of a forecasting task is to predict $L_{2}$ future timestamps using a sequence of $L_{1}$ preceding timestamps.
We employ a sliding window (of size $L$) technique, where $L=L_{1}+L_{2}$ to create MTS ready for the TSDE model. The specific dimensions for $L_{1}$ and $L_{2}$ are outlined in Table~\ref{tab:forecast_datasets}.

\begin{table}[h]
\caption[forecasting]{\small Description of Datasets for forecasting tasks. ``pre'': pretraining; ``ft'': finetuning.} 
\vspace{-10pt}
\label{tab:forecast_datasets}
\addtolength{\tabcolsep}{-1.8pt}
\centering
\footnotesize
\begin{tabular}{l |  c c c c c c c c c c c}
Dataset & \multicolumn{1}{c}{$K$}  & \multicolumn{1}{c}{$T$} & \multicolumn{1}{c}{$L_{1}$} & \multicolumn{1}{c}{$L_{2}$} & \multicolumn{1}{c}{$L$} & \multicolumn{1}{c}{$N_{\text{train}}$} & \multicolumn{1}{c}{$N_{\text{val}}$} & \multicolumn{1}{c}{$N_{\text{test}}$} & \multicolumn{1}{c}{$K_{\text{feat}}$} & \multicolumn{1}{c}{$E_{\text{pre}}$} & \multicolumn{1}{c}{$E_{\text{ft}}$} \\
\midrule
Electricity & 370 & 5,833 & 168 & 24 & 192 & 5,640 & 5 & 7 & 64 & 500 & 100\\
Solar & 137 & 7,009 & 168 & 24 & 192 & 6,816 & 5 & 7 & 128 & 50 & 25\\
Traffic & 963 & 4,001 & 168 & 24 & 192 & 3,808 & 5 & 7 & 128 & 400 & 90\\
Taxi & 1,214 & 1,488 & 48 & 24 & 72 & 1,415 & 5 & 56 & 128 & 400 & 20\\
Wiki & 2,000 & 792 & 90 & 30 & 120 & 701 & 5 & 5 & 128 & 400 & 20\\
\midrule
\end{tabular}
\vspace{-4pt}
\end{table}

\begin{table}[h]
\caption[]{\small Datasets specification for anomaly detection tasks. ``pre'': pretraining; ``ft'': finetuning.} 
\vspace{-10pt}
\label{tab:ping_ad_datasets}
\addtolength{\tabcolsep}{-3.6pt}
\centering
\footnotesize
\begin{tabular}{l c c | c c c | c c c c c c c c}
 &   & &  \multicolumn{3}{c|}{\# Timestamps} &  &  &  &  &  &  & \\
Dataset & \multicolumn{1}{c}{$K$}  &  \multicolumn{1}{c|}{$L$} & \multicolumn{1}{c}{Train} & \multicolumn{1}{c}{Val} & \multicolumn{1}{c|}{Test} & \multicolumn{1}{c}{$N_{\text{train}}$} & \multicolumn{1}{c}{$N_{\text{val}}$} & \multicolumn{1}{c}{$N_{\text{test}}$} & \multicolumn{1}{c}{$E_{\text{pre}}$} & \multicolumn{1}{c}{$E_{\text{ft}}$} & \multicolumn{1}{c}{$r$} & \multicolumn{1}{c}{seed} & \multicolumn{1}{c}{step}\\
\midrule
SMD & 38 & 100 & 566,724 & 141,681 & 708,420 & 5,667 & 1,416 & 7,084 & 250 & 30 & 0.5 & 42 & 100\\
MSL & 55 & 100 & 44,653 & 11,664 & 73,729 & 446 & 116 & 737 & 100 & 20 & 1 & 42 & 100\\
SMAP & 25 & 100 & 108,146 & 27,037 & 427,617 & 1,081 & 270 & 4,276 & 250 & 30 & 1 & 42 & 100\\
SWaT & 51 & 100 & 396,000 & 99,000 & 449,919 & 3,960 & 990 & 4,499 & 200 & 30 & 1 & 42 & 100\\
PSM & 25 & 100 & 105,984 & 26,497 & 87,841 & 1,059 & 264 & 878 & 50 & 10 & 1 & 42 & 100 \\
\midrule
\end{tabular}
\vspace{-2pt}
\end{table}
\subsubsection{Electricity}
The electricity dataset\footnote{ \url{https://archive.ics.uci.edu/dataset/321/electricityloaddiagrams20112014}}  \cite{misc_electricityloaddiagrams20112014_321} provides hourly electricity consumption of 370 customers. 
The training set features a single, long, and high-dimensional MTS, covering 370 customers across 5,833 timestamps. 
The testing set comprises 7 MTS, each with identical dimensions. 
While the customers are not necessarily identical across the different MTS, the time series for each customer belong to the same distribution, exhibiting a similar range of values. 
The training and validation sequences are created from the training set, while the testing sequences are derived from the test set. 
We use overlapping window with stride of 1 to create training MTS instances, and a sliding window with a stride equal to the prediction length for evaluation.
This guarantees that each evaluation in the validation and test sets is based on unique data segments.

\subsubsection{Solar}
The solar dataset\footnote{\href{https://www.nrel.gov/grid/solar-power-data.html}{https://www.nrel.gov/grid/solar-power-data.html}} provides hourly measurements of the photovoltaic production in 137 stations in Alabama. The training set consists of a single MTS, with 137 variables corresponding to the different stations and 7,009 timestamps for each. The testing set comprises 7 MTS of the same shape. Since the MTS in the test set are identical to the training MTS, we only use the training set to create training/validation/test splits. 
Training and validation sequences are generated from the initial 7009$-$24$\times$7 timestamps. 
The 7 testing sequences are formed using rolling windows with a stride matching the prediction length.

\subsubsection{Taxi}
The taxi dataset\footnote{\href{https://www.nyc.gov/site/tlc/about/tlc-trip-record-data.page}{https://www.nyc.gov/site/tlc/about/tlc-trip-record-data.page}} provides traffic time series of New York Taxis.
The rides values are recorded from 1,214 different locations with a frequency of 30 minutes for each location. The January 2015 data is used for training and January 2016 is used as the test set. Each MTS, in both the training and test sets, is of shape 1,214$\times$1,488 where 1,214 is the total number of the different features and 1,488 is the total number of timestamps for each time series. 
For training MTS instances, we employ overlapping windows with a stride of 1, and for non-overlapping evaluation, a sliding window with a stride set to the prediction length. This ensures that each evaluation in the validation and testing phases is conducted on distinct data segments. 
Following previous studies \cite{Nguyen_Quanz_2021, tashiro2021csdi}, we use 56 sequences to test TSDE's performance.

\subsubsection{Traffic}
The traffic dataset\footnote{\href{https://archive.ics.uci.edu/dataset/204/pems+sf}{https://archive.ics.uci.edu/dataset/204/pems+sf}} provides hourly occupancy rate, between 0 and 1, of 963 car lanes in the San Francisco Bay Area. 
In the training set, there is only one time series, encompassing 963 lanes with 4,001 timestamps for each. 
The testing set contains 7 MTS of the same shape. 
To create training MTS instances, we employ an overlapping window technique with a stride of 1. 
Conversely, for evaluation purposes in both validation and testing of MTS, we utilize a sliding window approach where the stride matches the prediction length, guaranteeing that each evaluation segment is unique and distinct. 
Following \cite{Nguyen_Quanz_2021, tashiro2021csdi}, we use the last 7 sequences of the test split as testing samples.

\subsubsection{Wiki}
The wiki dataset\footnote{\href{https://www.kaggle.com/c/web-traffic-time-series-forecasting/data}{https://www.kaggle.com/c/web-traffic-time-series-forecasting/data}}  consists of daily page views of about 145,000 Wikipedia pages. 
Salinas et al.~\cite{NEURIPS2019_0b105cf1} compiled a single time series for training from the original dataset, encompassing recordings from 9,013 pages over 792 timestamps. Additionally, they created a test set comprising 5 MTSs, each matching the training set in dimensionality.
Following \cite{tashiro2021csdi, Nguyen_Quanz_2021}, we use a subset of 2,000 pages to train and evaluate the models. 
We employ a sliding window to prepare the data for TSDE, with a stride of 1 for training MTS and a stride of prediction length for validation and testing.

\subsection{Anomaly Detection datasets}
Each dataset is preprocessed into a set of MTS using a non-overlapping sliding window of size 100, ensuring uniformity in the analysis. Below, we describe the five datasets employed in our experiments. The specifications of each dataset are summarized in Table ~\ref{tab:ping_ad_datasets}.

\subsubsection{Server Machine Dataset (SMD)} Introduced by \citet{10.1145/3292500.3330672}, the SMD is derived from a large Internet company's server data, spanning a duration of 5 weeks. This dataset encompasses 38 dimensions, capturing a wide range of metrics, like CPU load, network usage and memory usage, that are crucial for understanding server performance and anomalies. The observations in SMD are all equally-spaced 1 minute apart. The comprehensive nature of the SMD allows for a detailed analysis of server behavior under various conditions. There are about 4.16\% anomalous timestamps in the test set.

\subsubsection{Pooled Server Metrics (PSM)} Collected by \citet{10.1145/3447548.3467174}, the PSM dataset aggregates data from multiple application server nodes at eBay. With 26 dimensions, it provides a rich set of metrics that are instrumental in monitoring and detecting anomalies within eBay's server infrastructure. Similar to SMD, the features describe server machine metrics such as CPU utilization and memory. The training dataset spans 13 weeks, with an additional eight weeks designated for testing. Both the training and testing datasets contain anomalies. However, labels indicating these anomalies have been provided solely for the testing set after being meticulously crafted by engineers and domain experts. There are about 27.8\% anomalies in the test set. 

\subsubsection{Mars Science Laboratory (MSL)} MSL, a public contribution from NASA \cite{10.1145/3219819.3219845}, includes telemetry anomaly data derived from the Incident Surprise Anomaly (ISA) reports of spacecraft monitoring systems. It embodies 55 dimensions in the context of space exploration, where precision and reliability are paramount. This dataset has about 10.5\% anomalies in the test set.

\subsubsection{Soil Moisture Active Passive (SMAP)} SMAP, similar to SMD is provided by NASA \cite{10.1145/3219819.3219845}, includes telemetry anomaly data from the SMAP satellite with 25 dimension.  This dataset contains 12.8\% of timepoints that are anomalous in the test set.

\subsubsection{Secure Water Treatment (SWaT)} The SWaT dataset, provided by \citet{7469060}, is a testbed for attacks against industrial facilities. It includes data from 51 sensors over continuous operations in water treatment systems. The SWaT dataset allows for an in-depth exploration of attacks detection in environments where the stakes for security are exceptionally high. In this dataset, anomalies account for 12.1\% of the total timestamps.

\section{Metrics}
\begin{figure}[t]
  \begin{center}
    \includegraphics[width=0.4\textwidth]{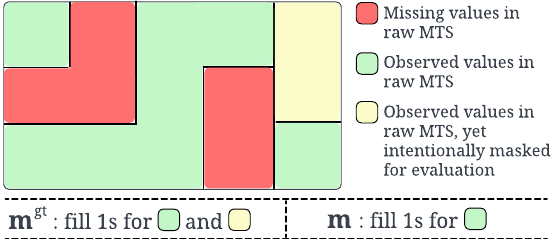}
  \end{center}
  \vspace{-6pt}
  \caption{Visualization of a MTS input $\mathbf{x}$, illustrating the formation of binary-valued evaluation masks $\mathbf{m}^{\text{gt}}$ and $\mathbf{m}$.}
  \label{fig:eval-mask-explain}
  \vspace{-2pt}
\end{figure}

In this section, we will elaborate all the metrics (with formula) that are used in our downstream tasks evaluation. 
In the scenario where missing/future values need to be predicted, i.e., imputation, interpolation and forecasting, we introduce a set of mathematical notations to precisely describe how the adopted metrics are computed. Let $\textbf{x}$ represent the MTS with ground truth values. 
To facilitate evaluation, we define two masks as visualized in Figure~\ref{fig:eval-mask-explain}:
\begin{itemize}[leftmargin=*]
    \item $\textbf{m}^{\text{gt}}$ marks the values that are originally available with 1s, including the ones used for evaluation, and the values missing in the original data with 0. It serves to identify the original ground truth observations within the data.
    \item $\textbf{m}$ represents the values that are known or observable to the model, excluding the masked values set aside for evaluation. This approach of selective masking enables an assessment of the model's performance on specific data segments.
\end{itemize}
Once all the missing values are imputed, we focus solely on the values that were specifically masked for evaluation. These are identified by the mask $\textbf{m}^{\text{eval}}=\textbf{m}^{\text{gt}}-\textbf{m} $. Given the probabilisitic nature of our model, we approximate the predicted distribution with 100 samples and denote it as $\hat{\textbf{F}}$. For deterministic evaluation, rather than relying on the entire distribution, we utilize the median of all samples, serving as a deterministic estimation (denoted by $\hat{\textbf{x}}$) of the missing values.  

Assuming our dataset has $N$ MTS instances, let $\textbf{X} \in \mathbb{R}^{N\times K\times L}$ be the set of all MTS $\textbf{x}$, and  $\hat{\textbf{X}} \in \mathbb{R}^{N\times K\times L}$ be the set of all predictions $\hat{\textbf{x}}$. For the metrics formulas in the upcoming sections, we will average their value over all $N$ samples.
\subsection{CRPS}
The CRPS (Continuous Ranked Probability Score) is a metric used to quantify the accuracy of probabilistic predictions. It compares an observed outcome to the predicted distribution, and measures their compatibility.
It is calculated by evaluating the cumulative distribution function (CDF) of the predicted probabilities and the CDF of the observed outcomes. 
The mean squared difference between these two CDF over all possible thresholds gives the CRPS value. 
A lower CRPS indicates a more accurate model, meaning the predicted probabilities are closer to the observed outcomes.
Formally, CRPS is defined as follows: 
\begin{equation}
CRPS(F,x)=\int_{\mathds{R}}(F(y)-\mathds{1}\{y>x\})^{2}dy,
\end{equation}
where $F$ is the the probability distribution and $x$ is the observed outcome.
To handle this metric, a parametrization of it is needed. Let's first define the Quantile Loss (QL):

\begin{equation}
QL_{p}(q,x)=(\mathds{1}\{y<q\}-p)(q-x).
\end{equation}
A quantile is the value below which a fraction of observations in a group fall. The previous loss at quantile level $\alpha$ can be expressed as:
\begin{equation}
\mathcal{L}_{\alpha}(q,x):=(\mathds{1}\{x<q\}-\alpha)(q-x), \text{where} \alpha \in [0,1].
\end{equation}
By setting $q=F^{-1}(\alpha)$, we can derive the following equation:
\begin{equation}
CRPS(F,x)=2\int_{0}^{1}\mathcal{L}_{\alpha}(F^{-1}(\alpha),x)d\alpha.
\end{equation}
The integral can be approximated by a discretized sum as follows:
\begin{equation}
CRPS(F,x)=\frac{2}{N}\sum_{i=1}^{N}\mathcal{L}_{\alpha_{i}}(F^{-1}(\alpha_{i}),x), \text{where}\, \alpha_{i}=i\times \gamma,
\end{equation}
where $\gamma$ refers to the tick value.  With these notations, $\gamma(N+1)=1$.

CRPS is a popular metric for probabilistic tasks, where understanding the uncertainty and range of possible outcomes is as crucial as predicting the most likely outcome as opposed to deterministic tasks.
In our experiments, we evaluate the performance on the values that are masked and imputed by TSDE, denoted by $m_{k,l}^{\text{eval}}=1$. We formulate the CRPS score obtained for a value $x_{k,l}$ as:

\begin{equation}
    CRPS_{k,l}=CRPS(\hat{f_{k,l}},x_{k,l}),
\end{equation}
where $\hat{f}_{k,l}$, represents the predicted distribution for the $x_{k,l}$ approximated by 100 predicted values.
In our setting, we compute CRPS using the average of the scores of all masked values for which a ground truth value exists, i.e.:
\begin{equation}
CRPS(\hat{\textbf{F}},\textbf{x})=\frac{\sum_{k,l}CRPS_{k,l} \times m_{k,l}^{\text{eval}}}{\sum_{k,l}|x_{k,l}| \times m_{k,l}^{\text{eval}}}.
\end{equation}

\subsection{CRPS-sum}
CRPS-sum is a metric commonly used for probabilistic forecasting. It refers to the CRPS for the distribution of the sum of all the features. The CRPS-sum is formulated as:
\begin{equation}
    CRPS-sum(\hat{\textbf{F}},\textbf{x})=\frac{\sum_{l}CRPS(\sum_{k}\hat{f_{k,l}},\sum_{k}x_{k,l})}{\sum_{k,l}|x_{k,l}|}.
\end{equation}
In our forecasting setup, for a MTS of dimensions $K \times L$, where $L=L_{1}+L_{2}$ with $L_{1}$ representing the historical window and $L_{2}$ the prediction window, TSDE imputes the values for the final $L_{2}$ timestamps. Subsequently, we assess TSDE's performance on these $L_{2}$ timestamps using the CRPS-sum metric, calculated as follows:

\begin{equation}
    CRPS-sum(\hat{\textbf{F}},\textbf{x})=\frac{\sum_{l=L-L_{2}}^{L}CRPS(\sum_{k}\hat{f_{k,l}},\sum_{k}x_{k,l})}{\sum_{l=L-L_{2}}^{L}\sum_{k}|x_{k,l}|}.
\end{equation}

\subsection{MAE}
MAE (Mean Absolute Error) is a metric to measure the error between predicted and true observations, which has the form of
\begin{equation}
    MAE(\hat{\textbf{X}},\textbf{X})=\frac{1}{N}\sum_{i=1}^{N}(\frac{1}{K\times L}\sum_{k,l}|\hat{x}_{i,k,l}-{x}_{i,k,l}|).
\end{equation}
In our case, as we are interested solely in the values that are masked for evaluation, we obtain MAE by averaging over all the values that were masked for evaluation. So, we formulate MAE as: 
\begin{equation}
MAE(\hat{\textbf{X}},\textbf{X})=\frac{\sum_{i=1}^{N}\sum_{k,l}|\hat{x}_{i,k,l}-{x}_{i,k,l}|\times m^{\text{eval}}_{i,k,l}}{\sum_{i=1}^{N}\sum_{k,l}m^{\text{eval}}_{i, k,l}}.
\end{equation}
\subsection{MSE}
MSE (Mean Square Error) is a metric for measuring the standard deviation of the residuals.  
Mathematically the MSE is denoted as:
\begin{equation}
    MSE(\hat{\textbf{X}},\textbf{X})=\frac{1}{N}\sum_{i=1}^{N}(\frac{1}{K\times L}\sum_{k,l}(\hat{x}_{i,k,l}-x_{i,k,l})^2).
\end{equation}
Again, as we only evaluate on the masked positions, the MSE formula becomes:
\begin{equation}
MSE(\hat{\textbf{X}},\textbf{X})=\frac{\sum_{i=1}^{N}\sum_{k,l}(\hat{x}_{i,k,l}-{x}_{i,k,l})^2\times m^{\text{eval}}_{i,k,l}}{\sum_{i=1}^{N}\sum_{k,l}m^{\text{eval}}_{i,k,l}}.
\end{equation}

\subsection{RMSE}
Similar to MSE, RMSE (Root Mean Square Error) is a metric used to measure the standard deviation of the residuals, in the same units as the target. 
Mathematically the RMSE is denoted as:
\begin{equation}
    RMSE(\hat{\textbf{X}},\textbf{X})=\sqrt{\frac{1}{N}\sum_{i=1}^{N}(\frac{1}{K\times L}\sum_{k,l}(\hat{x}_{i,k,l}-x_{i,k,l})^2)}.
\end{equation}
In our evaluation, we calculate the RMSE scores averaged over the values designated for assessment through the formula:
\begin{equation}
RMSE(\hat{\textbf{X}},\textbf{X})=\sqrt{\frac{\sum_{i=1}^{N}\sum_{k,l}(\hat{x}_{i,k,l}-{x}_{i,k,l})^2\times m^{\text{eval}}_{i,k,l}}{\sum_{i=1}^{N}\sum_{k,l}m^{\text{eval}}_{i,k,l}}}.
\end{equation}

\subsection{AUROC}
AUROC (Area Under Receiving Operating Characteristic Curve) is a fundamental metric for assessing binary classification models. It represents the model's ability to differentiate between the classes (positive/negative) over various threshold levels, especially when dealing with imbalanced datasets. By plotting the True Positive Rate (TPR) against the False Positive Rate (FPR), it illustrates the model's discrimination power. The AUROC measures the area under this curve and scales between 0 and 1, with a higher value indicating superior model performance. 
In our evaluation, we leverage AUROC metric as implemented in the scikit-learn \cite{scikit-learn} library.

\subsection{Precision}
Precision quantifies the accuracy of the positive predictions made by the model, defined as the ratio of true positives (TP) to the total number of predicted positives, which includes both true positives and false positives (FP). Mathematically, precision is expressed as:
\begin{equation}
Precision = TP/(TP+FP).
\end{equation}
We utilize the implementation from scikit-learn library \cite{scikit-learn}.
\subsection{Recall}
Recall, also known as Sensitivity, quantifies how well the model can identify the actual positives. It is defined as the ratio of TP to the sum of TP and false negatives (FN). The formula of recall is:
\begin{equation}
Recall = TP/(TP+FN).
\end{equation}
As for anomaly detection, missing an anomalous timestamp carries significant consequences, recall becomes a relevant metric.

\subsection{F1-score}
F1-score combines Recall and Precision and provides a balanced model performance accounting for the trade-offs between the two aforementioned metrics. F1-score is defined as the harmonic mean between recall and precision, giving equal weight to both of them, and penalizing extreme differences between them.
\begin{equation}
F1 = 2 \times \frac{Precision\times Recall}{Precision + Recall}.
\end{equation}
For anomaly detection experiments, we adopt the implementation from scikit-learn \cite{scikit-learn} library.

\subsection{RI}
RI (Rand Index) is a clustering metric used to evaluate similarity between two clustering assignments, regardless of the absolute associated labels. 
It is calculated by examining the agreement and disagreement of pairings within the clustering assignments relative to the true labels.
RI takes into account all sample pairs, counting those that are consistently grouped in the same or different clusters in both the clustering assignments and the true labels. This includes the number of agreeing pairs assigned to the same cluster and those correctly separated into distinct clusters.
\begin{equation}
RI = \frac{\text{Number of agreeing pairs}}{\text{Total number of pairs}}.
\end{equation}
The RI value ranges from 0 to 1, with a higher value indicating high similarity of the clustering to the true labels. 
We use the scikit-learn\cite{scikit-learn} implementation of RI in our experiments.

\subsection{Adjusted RI (ARI)}
The raw RI is adjusted to account for chance grouping and ensure that random labeling will have a very low score (close to 0) independently of cluster assignments. 
The ARI is a more accurate clustering similarity metric, with values ranging from -1 to 1. 
Values close to 1 suggest a more accurate cluster assignment. 
Negative values indicate a clustering result that is worse than a random assignment. 
\begin{equation}
ARI = \frac{\text{RI}-\text{Expected RI}}{\text{Max RI}-\text{Expected RI}},
\end{equation}
where Expected RI is the expected value of RI if the clusters are assigned randomly and Max RI is the maximum possible value of RI.
Scikit-learn\cite{scikit-learn} implementation is adopted.

\begin{table}[t!]
\centering
\caption{Summary of main layers specifications}
\vspace{-8pt}
\label{tab:model_layers}
\small
\begin{tabular}{l|l|l}

\textbf{Layer (type)}        & \textbf{Output Shape} & \textbf{Param \#}  \\ 
\hline
\text{embed\_layer (Embedding)}      & \text{[K, 16]}             & $K \times 16$  \\ 
\hline
conv (Linear)                & [*, K]             & $32 \times K \times K + K$  \\ 
\hline
mlp\_1 (Linear)              & [*, 256]           & $L \times K \times 33 \times 256 + 256$     \\ 
mlp\_1\_activation (SiLU)    & [*, 256]           & 0                          \\ 
mlp\_1\_dropout (Dropout)    & [*, 256]           & 0                         \\ 
mlp\_2 (Linear)              & [*, 256]           & $256 \times 256 + 256$     \\ 
mlp\_2\_activation (SiLU)    & [*, 256]           & 0                             \\ 
mlp\_2\_dropout (Dropout)    & [*, 256]           & 0                         \\
mlp\_3 (Linear)              & [*, C] & $256 \times C + C$ \\ 
\hline
\end{tabular}
\begin{flushleft}
\footnotesize
\ * indicates the batch size  
\end{flushleft}
\vspace{-2pt}
\end{table}

\section{Implementation}
In this section we delve into the architecture components and the layers that underpin our framework implementation. Our architecture is built on two key elements: an embedding function for processing observed segments and a denoising block for imputing the missing or the masked values, both implemented as Pytorch NN modules. Below, we provide more details about the the type of layers and the design of these components. Table~\ref{tab:model_layers} summarizes the main layers introduced in the subsequent sections.
\subsection{Embedding Function}
The embedding function stands as the pivotal component of TSDE, enabling us to adeptly handle a wide range of tasks by using the generated embeddings. This block combines four inputs and use them to transform the MTS into useful and generic representations: 
\begin{enumerate}[label={(\arabic*)},leftmargin=*]

    \item Time embedding $\textbf{s}_{\text{time}}$: We use 128-dimensional fixed embedding for each timestamp following \cite{tashiro2021csdi,zuo2020transformer} to encapsulate temporal information. Refer to \eqref{eq:temp_embedding} for a detailed formulation.
    
    \item Feature embedding $\textbf{s}_{\text{feat}}$: We leverage a Pytorch embedding layer to derive 16-demensional embeddings for each feature ID. 
    These weights of the embedding layer are trained, enabling the model to better understand and incorporate the non-temporal dependencies in the MTS embedding.

    \item Mask $\mathbf{m}^{\text{IIF}}$: A mask created using Algorithm~\ref{algo:masking}, indicating the observable segments.

    \item $\mathbf{x}_{0}^{\text{obs}}$: the observed values in the MTS, obtained by applying $\mathbf{m}^{\text{IIF}}$ to raw MTS $\mathbf{x}_{0}$, as formulated in \eqref{eq:obs_msk_parts}.
    
\end{enumerate}

The core of the embedding block invloves the Transfromer architecture. Our model utilizes dual-orthogonal Transformer encoders with a crossover mechanism, each realized with a single-layer TransformerEncoder as implemented in Pytorch\cite{NEURIPS2019_bdbca288}. 
This setup includes a multihead (=8) attention layer and incorporates fully connected layers and layer normalization. 
The input to the transformer encoders is a concatenation of $\textbf{s}_{\text{time}}$, $\textbf{s}_{\text{feat}}$ and $\textbf{x}_{0}^{\text{obs}}$. 
To align with the input requirement of multihead attention, $\textbf{x}_{0}^{\text{obs}}$ is first projected into a matrix of shape $K\times L\times 16$, ensuring that the last dimension of the concatenated tensor is divisible by 8.

The outputs of the two encoders, are projected to lower dimensional space and concatenated along with the $\textbf{m}^{\text{IIF}}$ mask, resulting in a refined embedding of the MTS (the observed segment).

\subsection{Denoising Block}
At diffusion step $t$, the denoising block (i.e.,~conditional reverse diffusion block) receives a set of inputs that are used for the conditioned denoising.
These inputs include the MTS embeddings from the embedding function, the masked and noisy segment of the original MTS at the $t$-th diffusion step, the diffusion step embedding, and a mask delineating the locations of added noise corresponding to the optimizable areas.
To prepare $\textbf{x}_{t}^{\text{msk}}$ for the denoising block, we derive it from $\textbf{x}_{0}^{\text{msk}}$ by

\begin{equation}
\mathbf{\textbf{x}}_{t}^{\text{msk}}=\sqrt{\tilde{\alpha_{t}}}\mathbf{x}_{0}^{\text{msk}}+\sqrt{1-\tilde{\alpha_{t}}}\boldsymbol{\epsilon},
\end{equation}
where $\tilde{\alpha_{t}}$ indicates noise level at diffusion step $t$, defined as 
\begin{equation}
     \tilde{\alpha_{t}}:=\textstyle{\prod}_{i=1}^{t} (1-\beta_{i}).
\end{equation}
Following \cite{tashiro2021csdi}, we set the total number of diffusion steps to $T$=50, and we use a quadratic noise scheduler formulated as: 
\begin{equation}
\beta_{t} = (\frac{T-t}{T+t}\sqrt{\beta_{1}}+\frac{t-1}{T-1}\sqrt{\beta_{T}})^2.
\end{equation}
The minimum noise level and the maximum noise level are set to $\beta_{1}=0.0001$ and $\beta_{T}=0.5$ respectively as in \cite{tashiro2021csdi}.

The denoising block is composed mainly of $Conv1\times 1$ layers, implemented using the Linear layer in Pytorch.
Conditioned on the embeddings generated by the embedding block, the denoising block plays a dual role that extends beyond mere denoising.
It encourages the embedding function to learn robust and meaningful representation of the MTS and updates this block weights during the training. 
Furthermore, in the context of filling in missing values, the denoising block plays a pivotal role in imputing missing values by filling them first with random values and then denoising them iteratively starting from diffusion step $t$=$T$=50 down to step t=1. 
During inference, the goal is to impute all the missing values (not only the masked ones for evaluation), hence $\textbf{x}_{0}^{\text{msk}}=(\mathds{1}_{K \times L}-\textbf{m})\odot \mathbf{x}_{0}$. 
A random noise is then injected into $\textbf{x}_{0}^{\text{msk}}$ to obtain $\textbf{x}_{T}^{\text{msk}}$; and TSDE denoise it resulting in imputed MTS, as detailed in Algorithm~\ref{algo:inference}.
Implementation-wise, we use a 4-layer residual module (we set the residual channels to 64), composed mainly of $\text{Conv}1\!\times\!1$ layers.

\begin{algorithm}[h!]
\footnotesize
\DontPrintSemicolon
\KwInput{Ground-truth MTS sample $\textbf{x}_{0}$, the trained denoising and embedding functions (approx. by NN): $\boldsymbol{\epsilon}_{\boldsymbol{\theta}}(\cdot)$ and $\textit{\textbf{f}}_{\boldsymbol{\phi}}(\cdot)$}
\KwOutput{The imputed MTS ${\textbf{x}}_{0}^{\text{msk}}$}

\For{$(t = T; t=1; t\minus \minus$)}{

    Obtain $\mathbf{m}$ denoting available values with 1 in $\textbf{x}_{0}$;

    $\small\smash{\mathbf{x}_0^{\text{obs}}}\gets \textbf{x}_{0}$ ;

    Sample a noise matrix $\boldsymbol{\epsilon} \sim \mathcal{N}(\mathbf{0},\textbf{I})$ that has the same shape as $\small\smash{\textbf{x}_{0}^{\text{obs}}}$;

    $\small\smash{\mathbf{x}_t^{\text{msk}}}\gets (1-m)\boldsymbol{\epsilon}$
    
    $\small\smash{\mathbf{x}_{t-1}^{\text{msk}}}\gets \boldsymbol{\mu}_{\boldsymbol{\theta}}(\textbf{x}_{t}^{\text{msk}}\!,t,\textit{\textbf{f}}_{\boldsymbol{\phi}}(\textbf{x}_{0}^{\text{obs}})) + \sigma_{t}.\epsilon$, cf.~\eqref{eq:condition_reverse_process} 
}

\textbf{return} $\small\smash{\textbf{x}_{0}^{\text{msk}}}$;
\caption{TSDE Inference Procedure}
\label{algo:inference}
\end{algorithm}

\subsection{Downstream-specific masks}

To tailor the model for specific missing values patterns, we finetune TSDE by employing task-specific masks.

\subsubsection{Imputation mask}
We randomly mask a ratio $r$ of the observed values to simulate missing data. The value of $r$ is sampled from the range [0.1,0.9] to cover different missing cases, yet keeping some observed values for conditionning the denoising process. This mask is useful when the missing values do not intricate some specific patterns within the data.

\setlength{\textfloatsep}{10pt plus 1.0pt minus 2.0pt}
\begin{algorithm}[h]
\footnotesize
\DontPrintSemicolon
\KwInput{Mask $\mathbf{m}\!=\!\{m_{1:K,1:L}\}\!\in\!\{0,1\}^{K\times L}$ indicating the missing values in $\mathbf{x}_0$}
\KwOutput{A pseudo observation mask $\mathbf{m}^{\text{imp}}\in\!\{0,1\}^{K\times L}$}


$r \gets$ random value from the range of [0.1, 0.9]; \tcp*{\scriptsize imputation mask ratio}

$N \gets \sum_{k=1}^K \sum_{l=1}^L m_{k,l}$; \tcp*{\scriptsize total number of observed values}


$\mathbf{m}^{\text{imp}} \gets \mathbf{m}$ and randomly set $\nint{N\times r}$ 1s to 0; \tcp*{\scriptsize apply imputation mask}

\textbf{return} $\mathbf{m}^{\text{imp}}$;
\caption{Imputation Mask}
\label{algo:imputation_masking}
\end{algorithm}

\setlength{\textfloatsep}{10pt plus 1.0pt minus 2.0pt}
\begin{algorithm}[t]
\footnotesize
\DontPrintSemicolon
\KwInput{Mask $\mathbf{m}\!=\!\{m_{1:K,1:L}\}\!\in\!\{0,1\}^{K\times L}$ indicating the missing values in $\mathbf{x}_0$, and Dataset $\mathcal{D}$ of all MTS samples}
\KwOutput{A pseudo observation mask $\mathbf{m}^{\text{hist}}\in\!\{0,1\}^{K\times L}$}

$p \gets$ random value from the range of [0, 1]; 

Draw $\tilde{\mathbf{x}_0} \sim \mathcal{D}$;

$\tilde{\mathbf{m}}\!=\!\{\tilde{m}_{1:K,1:L}\}\!\in\!\{0,1\}^{K\times L}$ indicating the missing values in $\tilde{\mathbf{x}}_0$;

$\mathbf{m}^{\text{hist}} \gets \tilde{\mathbf{m}}\odot \mathbf{m}$;

\uIf{$p>0.5$}{
    $\mathbf{m}^{\text{hist}} \gets$ imputation mask of m; \tcp*{\scriptsize Algorithm~\ref{algo:imputation_masking}}
}

\textbf{return} $\mathbf{m}^{\text{hist}}$;
\caption{History Mask}
\label{algo:history_mask}
\end{algorithm}

\setlength{\textfloatsep}{10pt plus 1.0pt minus 2.0pt}
\begin{algorithm}[t]
\footnotesize
\DontPrintSemicolon
\KwInput{Mask $\mathbf{m}\!=\!\{m_{1:K,1:L}\}\!\in\!\{0,1\}^{K\times L}$ indicating the missing values in $\mathbf{x}_0$}
\KwOutput{A pseudo observation mask $\mathbf{m}^{\text{int}}\in\!\{0,1\}^{K\times L}$}

$\mathbf{m}^{\text{int}} \gets \mathbf{m}$

$l' \gets$ uniformly sample a time step from $\mathbb{Z}\cap[1,L]$; \\
$\mathbf{m}^{\text{int}}[ : ,l'] \gets 0$; \tcp*{\scriptsize apply interpolation mask}

\textbf{return} $\mathbf{m}^{\text{int}}$;
\caption{Interpolation Mask}
\label{algo:interpolation_masking}
\end{algorithm}

\subsubsection{History mask}
The history mask combines the imputation mask or random masking strategy with a strategic approach exploiting the structured missing patterns observed in the data for improved imputation.
This is achieved by intersecting the observed indices of a given sample with the missing values from another randomly chosen sample to create a history mask.
Then, based on a sampled probability, we decide to either retain the history mask or apply the imputation mask, thereby effectively addressing the structured missing patterns.

\subsubsection{Interpolation mask}
This mask targets the MTS interpolation task. It operates by randomly selecting one timestamp and masking all values corresponding to the chosen timestamp.

\subsubsection{Forecasting mask}
Here, our approach involves masking a fixed number of timestamps at the end of the MTS. Specifically we mask the last $L_{2}$ timestamps of the MTS, where $L_{2}$ is the prediction window. This strategy simulates the forecasting scenario where the number of future values to predict is preset.
\begin{algorithm}[ht!]
\footnotesize
\DontPrintSemicolon
\KwInput{Mask $\mathbf{m}\!=\!\{m_{1:K,1:L}\}\!\in\!\{0,1\}^{K\times L}$ indicating the missing values in $\mathbf{x}_0$ and $L_{2}$ prediction window length}

\KwOutput{A pseudo observation mask $\mathbf{m}^{\text{for}}\in\!\{0,1\}^{K\times L}$}

$\mathbf{m}^{\text{for}} \gets \mathbf{m}$

$\mathbf{m}^{\text{for}}[ : ,-L_{2}:] \gets 0$; \tcp*{\scriptsize apply forecasting mask}

\textbf{return} $\mathbf{m}^{\text{for}}$;
\caption{Forecasting Mask}
\label{algo:forecasting_masking}
\end{algorithm}

\subsection{Projection Head for Anomaly Detection}
To enable anomaly detection with TSDE framework, we employ a dedicated projection layer to reconstruct the MTS from their embeddings. 
This reconstruction helps identifying deviations from normal patterns, which signify anomalies. The core of this projection is a Linear layer, implemented in Pytorch, designed to project the reshaped embeddings, excluding the mask  $\mathbf{m}^{\text{IIF}}$, from a shape of $(K\times 32, L)$ back to $(K, L)$. 
The finetuning of this projection head optimizes a MSE loss. This training objective is instrumental in guiding the projection layer to minimize the discrepancy between reconstructed and original MTS if there are no outliers. In our implementation, we used the Pytorch built-in MSE loss.   
\subsection{Classification}
For the classification task, we integrated a classifier head composed of MLP layer. The classifier is designed to transform the embeddings into class probabilities.
\begin{itemize}[leftmargin=*]
\item $1^{\text{st}}$ layer: consists of a fully connected layer followed by a SiLu activation function and a dropout layer. The linear layer project the flattened embeddings from $K\times L\times 33$ to 256, the SiLu function introduce non-linearity and the dropout layer randomly omits some of the layer outputs to prevent overfitting.
\item $2^{\text{nd}}$ layer: Similar to the first layer, with a fully connected layer connecting 256 nodes to 256 nodes followed by SiLu and dropout. It allows for increasing the model capacity and thus further extracting complex relationships from the embeddings and introduce more non-linear operation within the network.
\item $3^{\text{rd}}$ layer: is a simple linear projection layer, mapping the transformed features from 256 dimensions to the number of output classes (e.g., to 2 for binary classification).
\end{itemize}

\section{Experimental Setup}
This section delineates the comprehensive setup and configuration of our experiments. All experiments are executed on NVIDIA Tesla V100 GPUs. We set the random seed to 1 for all experiments except for anomaly detection, where we used a seed of 42.

\vspace{-4pt}
\subsection{Imputation}
\label{sec:exp_setup_imputation}
\subsubsection{Imputation on PhysioNet}
\label{sec:exp_setup_imputation_physio}
For the PhysioNet dataset, we set the batch size to 16. The model is pretrained for 2,000 epochs, followed by a finetuning phase of 200 epochs. During finetuning, we update all model weights, including the embeddings and denoising diffusion blocks, using the imputation mask (cf.~Algorithm~\ref{algo:imputation_masking}). The Adam optimizer, available in Pytorch, is employed throughout the pretraining and finetuning, starting with a learning rate of 0.001. This rate is scheduled to decay to 0.0001 at the $1,500^{\text{th}}$ epoch and further reduced to 0.00001 at $1,800^{\text{th}}$ epoch during pretraining. Similarly, during finetuning, the learning rate is reduced at the $150^{\text{th}}$ and $180^{\text{th}}$ epochs. This decay aligns precisely with a scheduled decay at 75\% and 90\% of the training total epochs, respectively for both pretraining and finetuning as in \cite{tashiro2021csdi}.

\vspace{-2pt}
\subsubsection{Imputation on PM2.5}
We maintain the same batch size of 16. The pretraining phase is shortened to 1,500 epochs, with finetuning further reduced to 100 epochs, reflecting the specific requirements and complexity of this dataset. Finetuning for this dataset entails an update of all model weights by employing a history mask (cf.~Algorithm~\ref{algo:history_mask}) tailored for the type of this dataset, having structured missing patterns, as opposed to IIF masking and imputation mask. We follow the same optimization setup using Adam optimizer with a starting learning rate of 0.001, systematically decaying to 0.0001 and 0.00001 at the 75\% and 90\% training completion respectively, during both pretraining and finetuning.

\vspace{-4pt}
\subsection{Interpolation}
PhysioNet is used for interpolation experiments. 
The experimental setup for this task mirrors Section ~\ref{sec:exp_setup_imputation_physio}. 
We use the same batch size, number of pretraining and finetuning epochs and optimizer configuration. Finetuning for the interpolation task involves updating both blocks (embedding and denoising) weights, using the interpolation masking, as described in Algorithm~\ref{algo:interpolation_masking}.

\vspace{-4pt}
\subsection{Forecasting}
\subsubsection{Probabilistic Multivariate Forecasting}

For forecasting, our experimental setup mirrors the imputation framework, using similar model hyperparameters and the same optimizer. 
However, to accommodate the high dimensionality in the forecasting MTS, we adjust the batch size to 8. Finetuning is carried out using the forecasting mask, as delineated in Algorithm~\ref{algo:forecasting_masking}, masking all the future values. 
The total number of pretraining and finetuning epochs for each dataset are presented in Table~\ref{tab:forecast_datasets}.

Given the extensive feature space of the five forecasting datasets, we adopt a feature subset sampling strategy during training, following \cite{tashiro2021csdi}. During training, a random subset of features is selected for each MTS instance within a batch, allowing for effective handling of the computational complexity of the attention mechanism. The number of sampled features is set to 64 for the Electricity dataset and 128 for the others as detailed in Table~\ref{tab:forecast_datasets}. The sampling during training results in selective updating and optimization of a subset of the feature embedding layer weights. By employing this sampling, an increase of the number of epochs is needed. The number of epochs was set based on the total number of features and the number of the sampled features.

\subsubsection{Deterministic multivariate forecasting}

For the determinitic multivariate forecasting benchmarking, five experiments were conducted on the preprocessed Electricity dataset provided by TimesNet authors~\cite{wu2023timesnet}. We used the same configuration and data splits as in~\cite{wu2023timesnet}. To create the MTS, we utilized a window of length $L = L_1 + L_2$, where $L_1$ is the history length and $l_2$ is the prediction length. The values employed for these lengths in our experiments for this benchmarking are: 8-8, 16-16, 32-32, 96-96, and 96-192 where the first value is $L_1$ and the second value is $L_2$. Each experiment was run only once.

The hyperparameters, such as batch size and the number of training epochs, are presented in Table~\ref{tab:tslib_benchmk_hyperparameters}. During training, we employed the IIF masking strategy. Larger prediction windows as in~\cite{wu2023timesnet}, could not be used with TSDE due to the out of memory limitation.

For all the other baselines from the time series library~\cite{wu2023timesnet}, we utilized the original implementations published by the authors, using the default hyperparameters provided in the experiments scripts to maintain consistenty with their original long-term forecasting experiments.

\begin{table}[h]
    \centering
    \caption{Hyperparameters used in the experiments for different window lengths}
    \label{tab:tslib_benchmk_hyperparameters}
    \centering
    \footnotesize
    \begin{tabular}{l|c|c|c|c|c}
        \textbf{Hyperparameters} & \multicolumn{1}{|c}{\textbf{8-8}} & \multicolumn{1}{|c}{\textbf{16-16}} & \multicolumn{1}{|c}{\textbf{32-32}} & \multicolumn{1}{|c}{\textbf{96-96}} & \multicolumn{1}{|c}{\textbf{96-192}} \\
        \midrule
        Batch size & 8 & 8 & 8 & 2 & 1 \\
        Number of training epochs & 100 & 100 & 100 & 20 & 20 \\
        \hline
    \end{tabular}
\end{table}

\subsection{Anomaly Detection}
For the anomaly detection we use the same pretraining setup with similar hyperparameters except for batch size and number of epochs. We set the batch size to 32, and adjust the total number of epochs for each dataset accordingly. We use a seed of 42 for all the anomaly detection experiments. Once TSDE is pretrained, we freeze all the model weights and finetune only the projection head weights. For finetuning, we used Adam optimizer in Pytorch and we set the learning rate to 0.0001 and the weight decay parameter to $10^{-6}$.

\subsubsection{Anomaly threshold setting}
Given the impracticality of accessing the size of the test subset in real-world scenarios, we adopt a pragmatic approach by fixing the anomaly detection threshold ($\delta$). This threshold is determined to ensure that the anomaly scores for r time points within the validation set exceed $\delta$, thereby classifying them as anomalies. This strategic threshold setting maintains consistency with the anomaly ratios observed in baseline methodologies\cite{xu2022anomaly}. In our experiments we used the same anomaly ratios employed by the baselines, as outlined in Table~\ref{tab:ping_ad_datasets}. 

\vspace{-2pt}
\subsubsection{Adjustement strategy}
TSDE, following \cite{zhou2023one, wu2023timesnet, xu2022anomaly}, incorporates a nuanced adjustment strategy for the detection of anomalies within successive abnormal segments. This strategy is predicated on the premise that the detection of a single time point as anomalous within a continuous abnormal segment warrants the classification of all time points within this segment as anomalies. This approach is derived from practical observations in real-world scenarios, where the identification of an anomaly triggers an alert, thereby bringing the entire abnormal segment to attention.

\vspace{-4pt}
\subsection{Classification}
To evaluate TSDE on the classification task, we perform two runs. In the first run, we pretrain the TSDE model specifically for the imputation task on the PhysioNet data with 10\% missing data ratio. This pretraining follows the experimental setup detailed in Section ~\ref{sec:exp_setup_imputation_physio}. Upon completing the pretraining phase, we proceed with inference on the training, validation, and test datasets, and save the median of the generated samples along with their respective labels. Subsequently, we freeze the weights of the embedding block and proceed to finetune only the classifier head of the pretrained model. This finetuning phase spans 40 epochs, during which the embedding block is fed with the imputed MTS. The primary training objective is minimizing the cross-entropy loss as implemented in PyTorch. To achieve this, we utilize the Adam optimizer, setting the learning rate to 0.0001 and applying a weight decay of $10^{-6}$. 

\vspace{-4pt}
\subsection{Clustering}
The clustering is tested under three settings: (1) raw MTS, (2) embedding of imputed MTS and (3) embedding of raw MTS. To generate the MTS embeddings, we employ the same pretrained model as used in the classification task. Each of the three aforementioned tensors is then projected into 2-dimensional space using UMAP with the Jaccard distance metric. For the clustering in this 2-dimensional projected space, we utilize the DBSCAN algorithm, as implemented in the scikit-learn library.

\section{Additional results}
Full results of comparing TSDE to recent baselines from the time series library~\cite{wu2023timesnet} are in Table~\ref{tab:tslib_full_forecasting_results}.
\begin{table*}[t]
  \caption{Full results for the forecasting task on Electricity following~\cite{wu2023timesnet} setting. We compare extensive competitive models under five different history-prediction lengths \{8-8, 16-16, 32-32, 96-96, 96-192\}. \emph{Avg} is averaged from all five history-prediction window lengths.}\label{tab:tslib_full_forecasting_results}
  \vskip 0.05in
  \centering
  \resizebox{1\textwidth}{!}{
  \begin{threeparttable}
  \begin{small}
  \renewcommand{\multirowsetup}{\centering}
  \setlength{\tabcolsep}{1pt}
  \begin{tabular}{c|c|cc|cc|cc|cc|cc|cc|cc|cc|cc|cc|cc|cc}
    \toprule
    \multicolumn{2}{c}{\multirow{2}{*}{Models}} &
    \multicolumn{2}{c}{\rotatebox{0}{\scalebox{0.8}{\textbf{TSDE}}}} & 
    \multicolumn{2}{c}{\rotatebox{0}{\scalebox{0.8}{TimesNet}}} &
    \multicolumn{2}{c}{\rotatebox{0}{\scalebox{0.8}{{ETSformer}}}} &
    \multicolumn{2}{c}{\rotatebox{0}{\scalebox{0.8}{LightTS$^\ast$}}} &
    \multicolumn{2}{c}{\rotatebox{0}{\scalebox{0.8}{DLinear$^\ast$}}} &
    \multicolumn{2}{c}{\rotatebox{0}{\scalebox{0.8}{FEDformer}}} & \multicolumn{2}{c}{\rotatebox{0}{\scalebox{0.8}{Stationary}}} & \multicolumn{2}{c}{\rotatebox{0}{\scalebox{0.8}{Autoformer}}} & \multicolumn{2}{c}{\rotatebox{0}{\scalebox{0.8}{Pyraformer}}} &  \multicolumn{2}{c}{\rotatebox{0}{\scalebox{0.8}{Informer}}} & \multicolumn{2}{c}{\rotatebox{0}{\scalebox{0.8}{Reformer}}}   & \multicolumn{2}{c}{\rotatebox{0}{\scalebox{0.8}{PatchTST}}} \\
    \multicolumn{2}{c}{} & \multicolumn{2}{c}{\scalebox{0.8}{(\textbf{Ours})}} & 
    \multicolumn{2}{c}{\scalebox{0.8}{\citeyearpar{wu2023timesnet}}} &
    \multicolumn{2}{c}{\scalebox{0.8}{\citeyearpar{woo2023etsformer}}} &
    \multicolumn{2}{c}{\scalebox{0.8}{\citeyearpar{zhang2022more}}} & \multicolumn{2}{c}{\scalebox{0.8}{\citeyearpar{Zeng2022AreTE}}} & \multicolumn{2}{c}{\scalebox{0.8}{\citeyearpar{pmlr-v162-zhou22g}}} & \multicolumn{2}{c}{\scalebox{0.8}{\citeyearpar{NEURIPS2022_4054556f}}} & \multicolumn{2}{c}{\scalebox{0.8}{\citeyearpar{NEURIPS2021_bcc0d400}}} &  \multicolumn{2}{c}{\scalebox{0.8}{\citeyearpar{liu2022pyraformer}}} & 
    \multicolumn{2}{c}{\scalebox{0.8}{\citeyearpar{Zhou_Zhang_Peng_Zhang_Li_Xiong_Zhang_2021}}}& \multicolumn{2}{c}{\scalebox{0.8}{\citeyearpar{kitaev2020reformer}}}& \multicolumn{2}{c}{\scalebox{0.8}{\citeyearpar{nie2023a}}}\\
    \cmidrule(lr){1-2}
    \cmidrule(lr){3-4} \cmidrule(lr){5-6}\cmidrule(lr){7-8} \cmidrule(lr){9-10}\cmidrule(lr){11-12}\cmidrule(lr){13-14}\cmidrule(lr){15-16}\cmidrule(lr){17-18}\cmidrule(lr){19-20}\cmidrule(lr){21-22}\cmidrule(lr){23-24}\cmidrule(lr){25-26}
    \multicolumn{2}{c}{Metric} & \scalebox{0.78}{MSE} & \scalebox{0.78}{MAE} & \scalebox{0.78}{MSE} & \scalebox{0.78}{MAE} & \scalebox{0.78}{MSE} & \scalebox{0.78}{MAE} & \scalebox{0.78}{MSE} & \scalebox{0.78}{MAE} & \scalebox{0.78}{MSE} & \scalebox{0.78}{MAE} & \scalebox{0.78}{MSE} & \scalebox{0.78}{MAE} & \scalebox{0.78}{MSE} & \scalebox{0.78}{MAE} & \scalebox{0.78}{MSE} & \scalebox{0.78}{MAE} & \scalebox{0.78}{MSE} & \scalebox{0.78}{MAE} & \scalebox{0.78}{MSE} & \scalebox{0.78}{MAE} & \scalebox{0.78}{MSE} & \scalebox{0.78}{MAE}  & \scalebox{0.78}{MSE} & \scalebox{0.78}{MAE} \\
    \toprule
    \multirow{5}{*}{\rotatebox{90}{\scalebox{0.95}{Electricity}}} 
    & \scalebox{0.78}{8-8} &\scalebox{0.78}{\textbf{0.189}} &\scalebox{0.78}{\textbf{0.288}} 
    &\scalebox{0.78}{0.318} &\scalebox{0.78}{0.380} &\scalebox{0.78}{0.207} &\scalebox{0.78}{0.333} &\scalebox{0.78}{0.292} &\scalebox{0.78}{0.391} &\scalebox{0.78}{0.803} &\scalebox{0.78}{0.719} &\scalebox{0.78}{0.205} &\scalebox{0.78}{0.327} &\scalebox{0.78}{0.303} &\scalebox{0.78}{0.361} &\scalebox{0.78}{0.198} &\scalebox{0.78}{0.321} &\scalebox{0.78}{0.228} &\scalebox{0.78}{0.228} &\scalebox{0.78}{0.228} &\scalebox{0.78}{0.338} 
    &\scalebox{0.78}{0.243} &\scalebox{0.78}{0.352} &\scalebox{0.78}{0.872} &\scalebox{0.78}{0.660}\\ 
    & \scalebox{0.78}{16-16} &\scalebox{0.78}{0.188} &\scalebox{0.78}{0.277} 
    &\scalebox{0.78}{\textbf{0.157}} &\scalebox{0.78}{\textbf{0.261}} &\scalebox{0.78}{0.237} &\scalebox{0.78}{0.360} &\scalebox{0.78}{0.221} &\scalebox{0.78}{0.323} &\scalebox{0.78}{0.357} &\scalebox{0.78}{0.425} &\scalebox{0.78}{0.180} &\scalebox{0.78}{0.300} &\scalebox{0.78}{0.149} &\scalebox{0.78}{0.255} &\scalebox{0.78}{0.170} &\scalebox{0.78}{0.294} &\scalebox{0.78}{0.241} 
    &\scalebox{0.78}{0.351} &\scalebox{0.78}{0.253} &\scalebox{0.78}{0.359} 
    &\scalebox{0.78}{0.269} &\scalebox{0.78}{0.369} &\scalebox{0.78}{0.270} &\scalebox{0.78}{0.332}\\
    & \scalebox{0.78}{32-32} &\scalebox{0.78}{0.158} &\scalebox{0.78}{\textbf{0.237}} 
    &\scalebox{0.78}{\textbf{0.147}} &\scalebox{0.78}{0.250}  &\scalebox{0.78}{0.245} &\scalebox{0.78}{0.363} &\scalebox{0.78}{0.189} &\scalebox{0.78}{0.291} &\scalebox{0.78}{0.243} &\scalebox{0.78}{0.330} &\scalebox{0.78}{0.186} &\scalebox{0.78}{0.301} &\scalebox{0.78}{0.152} &\scalebox{0.78}{0.254} &\scalebox{0.78}{0.170} &\scalebox{0.78}{0.290} &\scalebox{0.78}{0.260} &\scalebox{0.78}{0.364} &\scalebox{0.78}{0.310} &\scalebox{0.78}{0.403} 
    &\scalebox{0.78}{0.291} &\scalebox{0.78}{0.387} &\scalebox{0.78}{0.198} &\scalebox{0.78}{0.271}\\
    & \scalebox{0.78}{96-96} &\scalebox{0.78}{0.176} &\scalebox{0.78}{\textbf{0.247}} 
    &\scalebox{0.78}{\textbf{0.168}} &\scalebox{0.78}{0.272} &\scalebox{0.78}{0.249} &\scalebox{0.78}{0.353} &\scalebox{0.78}{0.213} &\scalebox{0.78}{0.316} &\scalebox{0.78}{0.197} &\scalebox{0.78}{0.282} &\scalebox{0.78}{0.193} &\scalebox{0.78}{0.308} &\scalebox{0.78}{0.170} &\scalebox{0.78}{0.271} &\scalebox{0.78}{0.201} &\scalebox{0.78}{0.317} &\scalebox{0.78}{0.386} &\scalebox{0.78}{0.449}&\scalebox{0.78}{0.323} &\scalebox{0.78}{0.409} 
    &\scalebox{0.78}{0.304} &\scalebox{0.78}{0.392} &\scalebox{0.78}{0.182} &\scalebox{0.78}{0.271}\\
    & \scalebox{0.78}{96-192} &\scalebox{0.78}{\textbf{0.132}} &\scalebox{0.78}{\textbf{0.218}} 
    &\scalebox{0.78}{0.184} &\scalebox{0.78}{0.289} &\scalebox{0.78}{0.269} &\scalebox{0.78}{0.365} &\scalebox{0.78}{0.221} &\scalebox{0.78}{0.325} &\scalebox{0.78}{0.196} &\scalebox{0.78}{0.285} &\scalebox{0.78}{0.201} &\scalebox{0.78}{0.315} &\scalebox{0.78}{0.182} &\scalebox{0.78}{0.283} &\scalebox{0.78}{0.222} &\scalebox{0.78}{0.334} &\scalebox{0.78}{0.378} &\scalebox{0.78}{0.443} &\scalebox{0.78}{0.360} &\scalebox{0.78}{0.440} 
    &\scalebox{0.78}{0.327} &\scalebox{0.78}{0.408} &\scalebox{0.78}{0.186} &\scalebox{0.78}{0.275}\\
    \cmidrule(lr){2-26}
    & \scalebox{0.78}{Avg} &\scalebox{0.78}{\boldres{0.169}} &\scalebox{0.78}{\boldres{0.253
}} 
    &\scalebox{0.78}{0.195} &\scalebox{0.78}{0.290} &\scalebox{0.78}{0.241} &\scalebox{0.78}{0.355} &\scalebox{0.78}{0.227} &\scalebox{0.78}{0.329} &\scalebox{0.78}{0.359} &\scalebox{0.78}{0.408}
    &\scalebox{0.78}{0.193} &\scalebox{0.78}{0.310} &\scalebox{0.78}{0.191} &\scalebox{0.78}{0.285} &\scalebox{0.78}{0.192} &\scalebox{0.78}{0.311} &\scalebox{0.78}{0.299} &\scalebox{0.78}{0.367} &\scalebox{0.78}{0.295} &\scalebox{0.78}{0.390} &\scalebox{0.78}{0.287} &\scalebox{0.78}{0.382} 
    &\scalebox{0.78}{0.342}&\scalebox{0.78}{0.362}\\
    \midrule
    \multicolumn{2}{c}{\scalebox{0.78}{{$1^{\text{st}}$ Count}}} & \multicolumn{2}{c}{\boldres{\scalebox{0.78}{\textbf{6}}}} & \multicolumn{2}{c}{\secondres{\scalebox{0.78}{4}}} & \multicolumn{2}{c}{\scalebox{0.78}{0}} & \multicolumn{2}{c}{\scalebox{0.78}{0}} & \multicolumn{2}{c}{\scalebox{0.78}{0}} & \multicolumn{2}{c}{\scalebox{0.78}{0}} & \multicolumn{2}{c}{\scalebox{0.78}{0}} & \multicolumn{2}{c}{\scalebox{0.78}{0}} &  \multicolumn{2}{c}{\scalebox{0.78}{0}} & \multicolumn{2}{c}{\scalebox{0.78}{0}} & \multicolumn{2}{c}{\scalebox{0.78}{0}} & \multicolumn{2}{c}{\scalebox{0.78}{0}} \\
    \bottomrule
  \end{tabular}
    \begin{tablenotes}
        \footnotesize
        \item[] $\ast$ means that there are some mismatches between our input-output setting and their papers. We adopt their official codes and only change the length of input and output sequences for a fair comparison.
  \end{tablenotes}
    \end{small}
  \end{threeparttable}
}
\end{table*}

\section{Key Findings and Limitations}
This work explored the integration of diffusion models and transformer encoders for TSRL. Our results indicate that conditioning the diffusion model on learned embeddings improves performance across tasks such as imputation, interpolation, and forecasting. Additionally, the embedding block within TSDE proves highly effective in handling sparse data by leveraging the SSL task of imputing missing values. The use of IIF masking facilitated the model's robustness to sparse data, and handling of different missingness scenarios.

Despite these advancements, the TSDE model has few limitations. The iterative nature of the denoising diffusion probablistic model can lead to slower inference, presenting a trade-off between enhanced quality and efficiency in real-world applications. Incorporating the SSL pretext task of IIF masking requires additional training epochs, which can be a constraint for rapid model deployment and retraining. Furthermore, while TSDE outperforms other methods, there remains a gap in perfectly matching the ground truth in highly noisy scenarios. Addressing these limitations will be crucial for further improving the model's robustness and applicability.
\section{Visualizations}
\label{appendix-vis}
\begin{figure*}[t!]
\centering
\includegraphics[height=21.5cm, width=16.5cm]{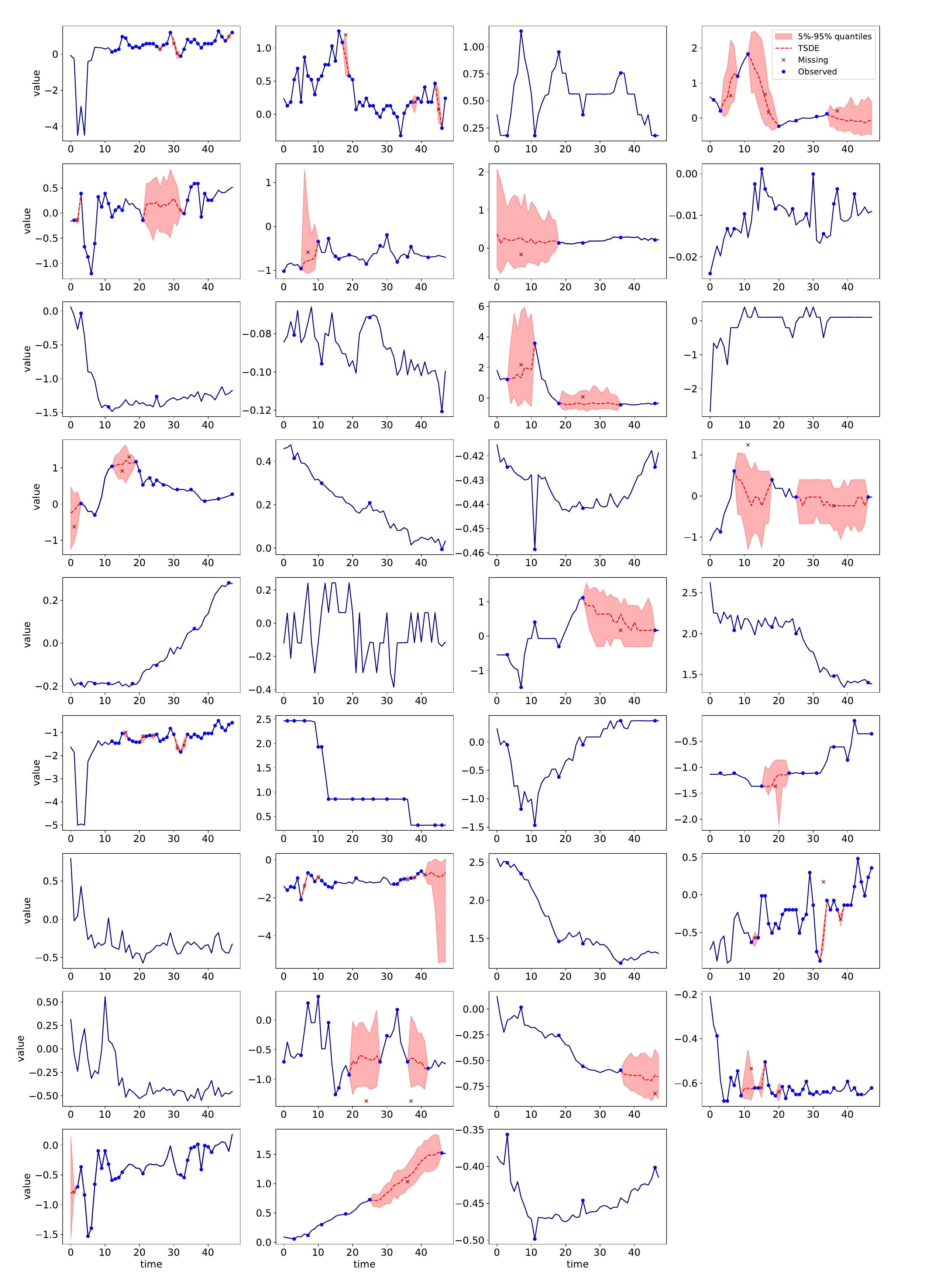}
\vspace{-15pt}
\caption{Prediction visualization for PhysioNet dataset, imputation task. The blue line and the red dashed line indicate the median of the generated samples. The red shade represents 5\%-95\% quantiles for the missing values.} \label{fig:imputation_viz_all_crop}
\vspace{-2pt}
\end{figure*}

\begin{figure*}[t!]
\centering
\includegraphics[height=21.5cm, width=16.5cm]{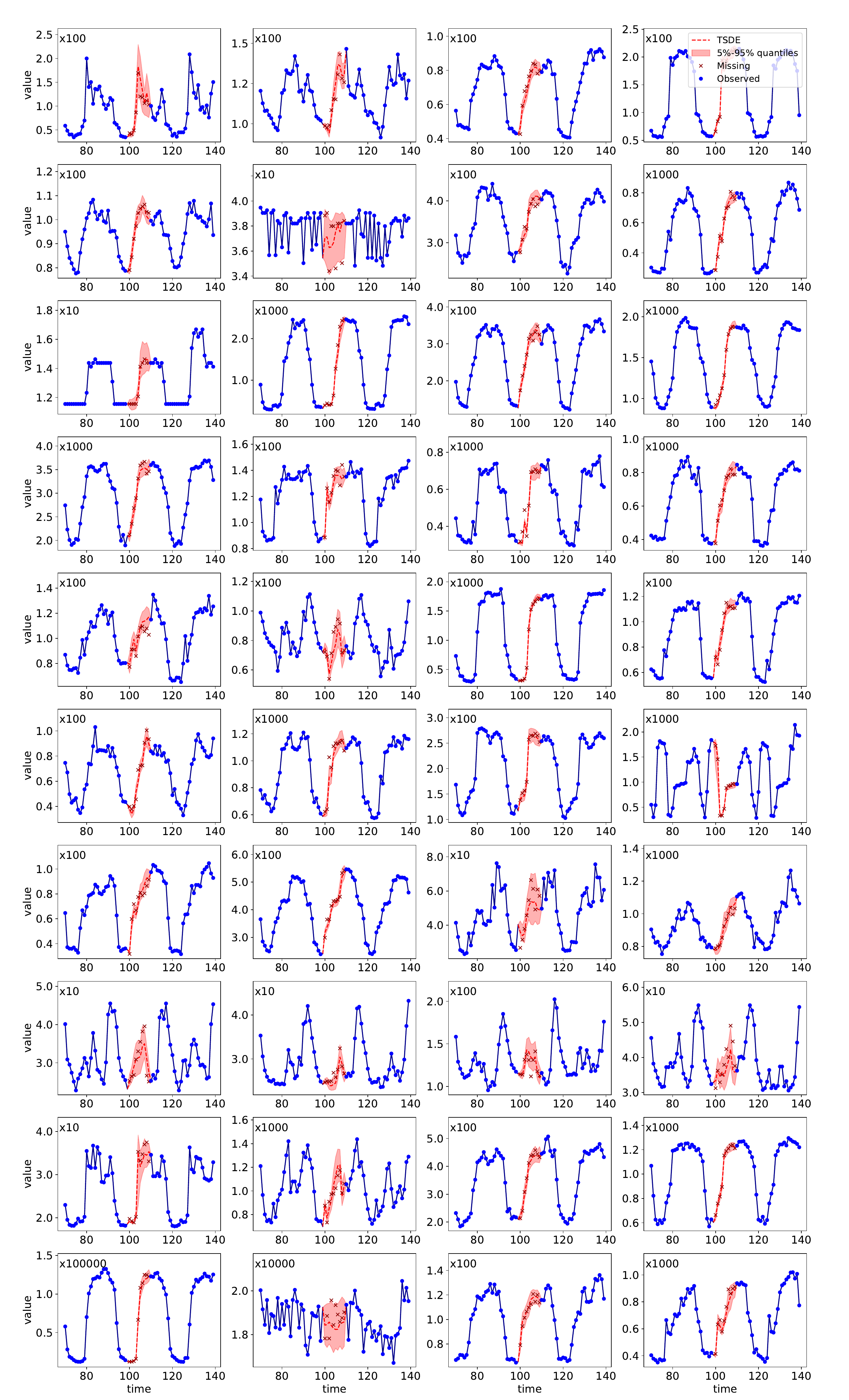}
\vspace{-15pt}
\caption{Prediction visualization for Electricity dataset, interpolation task. The blue line and the red dashed line indicate the median of the generated samples. The red shade represents 5\%-95\% quantiles for the missing values.} \label{fig:interpolation_viz_all_crop}
\vspace{-2pt}
\end{figure*}

\begin{figure*}[t!]
\centering
\includegraphics[height=21.5cm, width=16.5cm]{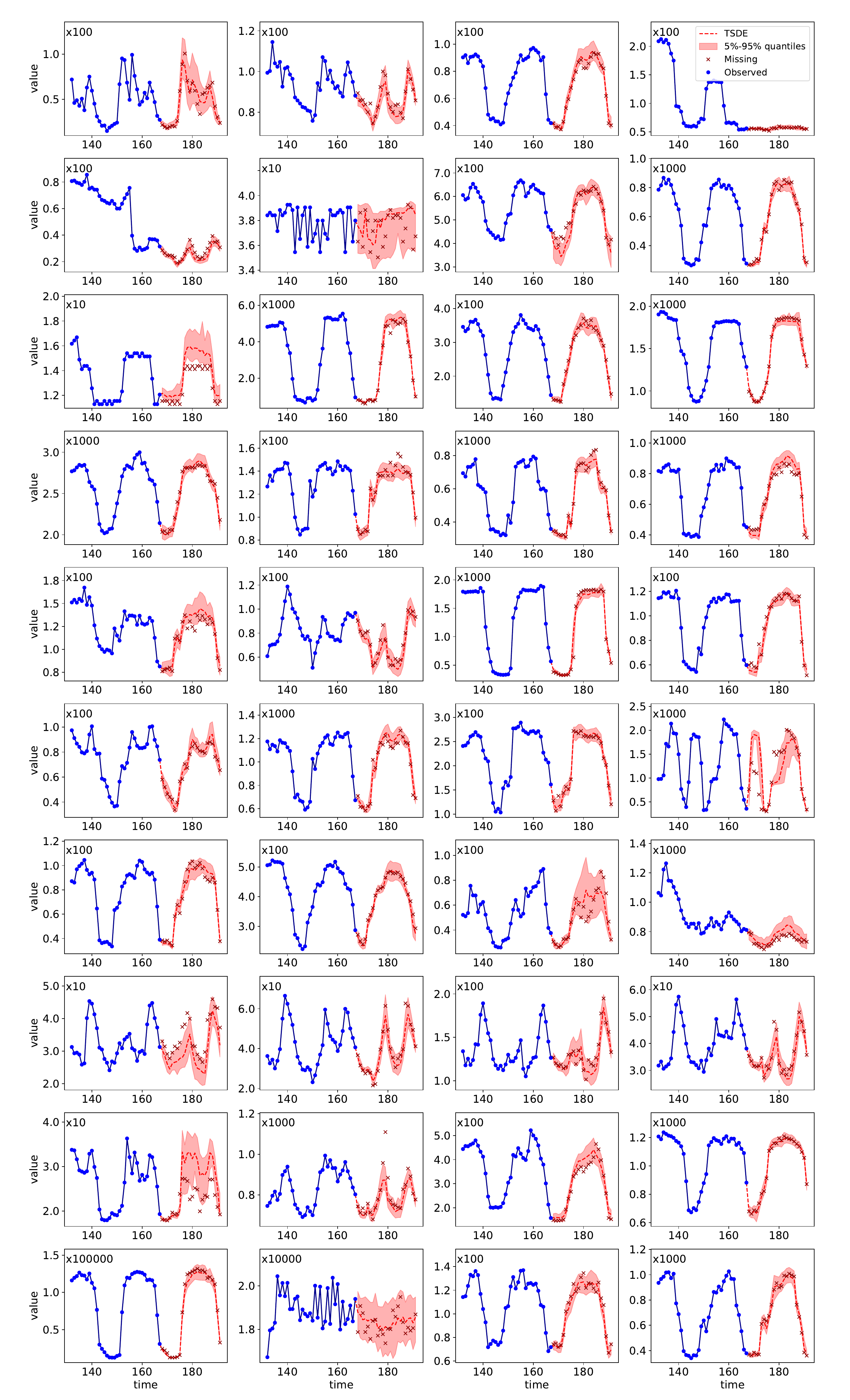}
\vspace{-15pt}
\caption{Prediction visualization for Electricity dataset, forecasting task. The blue line and the red dashed line indicate the median of the generated samples. The red shade represents 5\%-95\% quantiles for the forecasted future values.} \label{fig:forecasting_viz_all_crop}
\vspace{-2pt}
\end{figure*}

To showcase some qualitative results, we visualize experimental results from imputation, interpolation, and forecasting tasks in Figure~\ref{fig:imputation_viz_all_crop}, \ref{fig:interpolation_viz_all_crop} and \ref{fig:forecasting_viz_all_crop}, respectively.

\end{document}